\documentclass{article}

\usepackage[utf8]{inputenc} 
\usepackage{hyperref}       
\usepackage{url}            
\usepackage{booktabs}       
\usepackage{amsfonts}       
\usepackage{nicefrac}       
\usepackage{microtype}      
\usepackage{xcolor}         
\usepackage[pdftex]{graphicx}
\usepackage{verbatim} 
\usepackage{graphicx}
\usepackage{subfig}
\usepackage{pdflscape}
\usepackage{amsmath}
\usepackage[T1]{fontenc}    
\usepackage{authblk}

\usepackage{natbib}

\usepackage[preprint]{neurips_2021}

\title{Faking feature importance: A cautionary tale on the use of differentially-private synthetic data}

\author[1*]{Oscar Giles\thanks{Equal contribution. Listing order is alphabetical.}}
\author[1*]{Kasra Hosseini}
\author[1*]{Grigorios Mingas}
\author[1*]{Oliver Strickson}
\author[1]{Louise Bowler}
\author[1]{Camila Rangel Smith}
\author[2]{Harrison Wilde}
\author[2]{Jen Ning Lim}
\author[3]{Bilal Mateen}
\author[4]{Kasun Amarasinghe}
\author[4]{Rayid Ghani}
\author[5]{Alison Heppenstall}
\author[6]{Nik Lomax}
\author[6]{Nick Malleson}
\author[1]{Martin O'Reilly}
\author[7]{Sebastian Vollmer}

\affil[1]{The Alan Turing Institute, UK}
\affil[2]{University of Warwick, Coventry, UK}
\affil[3]{University College London, London, UK}
\affil[4]{Carnegie Mellon University, Pittsburgh, Pennsylvania, USA}
\affil[5]{University of Glasgow, Glasgow, UK}
\affil[6]{University of Leeds, Leeds, UK}
\affil[7]{DFKI, Kaiserslautern, Germany}

\begin{document}

\newpage

\maketitle

\begin{abstract}
Synthetic datasets are often presented as a silver-bullet solution to the problem of privacy-preserving data publishing. However, for many applications, synthetic data has been shown to have limited utility when used to train predictive models. One promising potential application of these data is in the exploratory phase of the machine learning workflow, which involves understanding, engineering and selecting features. This phase often involves considerable time, and depends on the availability of data. There would be substantial value in synthetic data that permitted these steps to be carried out while, for example, data access was being negotiated, or with fewer information governance restrictions. This paper presents an empirical analysis of the agreement between the feature importance obtained from raw and from synthetic data, on a range of artificially generated and real-world datasets (where feature importance represents how useful each feature is when predicting a the outcome). We employ two differentially-private methods to produce synthetic data, and apply various utility measures to quantify the agreement in feature importance as this varies with the level of privacy. Our results indicate that synthetic data can sometimes preserve several representations of the ranking of feature importance in simple settings but their performance is not consistent and depends upon a number of factors. Particular caution should be exercised in more nuanced real-world settings, where synthetic data can lead to differences in ranked feature importance that could alter key modelling decisions. This work has important implications for developing synthetic versions of highly sensitive data sets in fields such as finance and healthcare.
\end{abstract}

\section{Introduction}

Access to high-quality data is a key concern of every data scientist
and machine-learning practitioner.  When the use of a dataset is restricted, perhaps to respect the privacy of individuals, the ability to perform analyses on the data, share results, or publish the analysis for reproduction by others is often limited. Restrictions might also incur additional infrastructure costs. This is often the case when dealing with datasets in domains such as healthcare, finance, telecommunications and the public sector, where access is tightly controlled using secure compute environments \citep{hdr}. 

Methods for producing synthetic data promise to address these limitations, by offering data that: (i) has the same form as the
original data, (ii) preserves certain properties of the underlying dataset (such that it is has a degree of utility for its intended
application), and (iii) limits inferences about individual records in
the original dataset (thus preserving the privacy of individuals).
The latter property means that synthetic datasets can often be treated
as less sensitive, if not completely publicly disclosable.

The requirement of synthetic data to have the same form as the original data, (i), means that synthetic data can be
substituted for the data it intends to replace or augment.  This requirement sets it
apart from other data intended for release, such as aggregate statistics or
coarsened data, as well as other products of modelling, such as trained generative models themselves.  The property
(ii) we refer to generally as \emph{utility}, and (iii) as \emph{privacy}. 

In this paper we will also use the term \emph{dummy data} to describe synthetic data that satisfies properties (i) and (iii) only, which are useful for developing and validating the mechanics of a data science pipeline (e.g. a machine learning pipeline), but not for other purposes.

One might expect there to be a trade-off between the privacy and utility of
synthesis methods. The data of highest utility
possible, by any reasonable measure, is the original data. A synthesis method
cannot add new information to the dataset, but it may obfuscate or fail to
capture relationships present in the original data, resulting in a reduction of utility. On
the other hand, releasing the original data offers no enhancement to privacy,
whereas releasing no data at all---or synthetic data generated with no
reference to the original data---offers maximal privacy, but for most purposes, low utility.
This \emph{privacy--utility trade-off} is explored in a growing literature on the subject, see for example  \citet{stadler2020, karr2006, jordonPATEGANGeneratingSynthetic2019, wilde2021}.

Other anonymisation and pseudonimisation techniques (for example, aggregation, the addition of random noise to records, outlier removal, record permutation, subsampling) share a similar trade-off between utility and privacy, with various issues documented in the literature. For example, removing names and identifiers may not offer good privacy;
keeping the data at the level of individuals without aggregating is sometimes considered to be a high risk approach, for anything beyond a very small set of released features \citep{ico}; aggregation can destroy the utility of the data for many types of analysis \citep{drechsler2011, karr2006}; many techniques offer no theoretical privacy guarantees; re-identification attacks on supposedly anonymised data have been successful \citep{reidentification, reidentification2}.

\begin{figure*}
\centering
\includegraphics[width=1.0\linewidth]{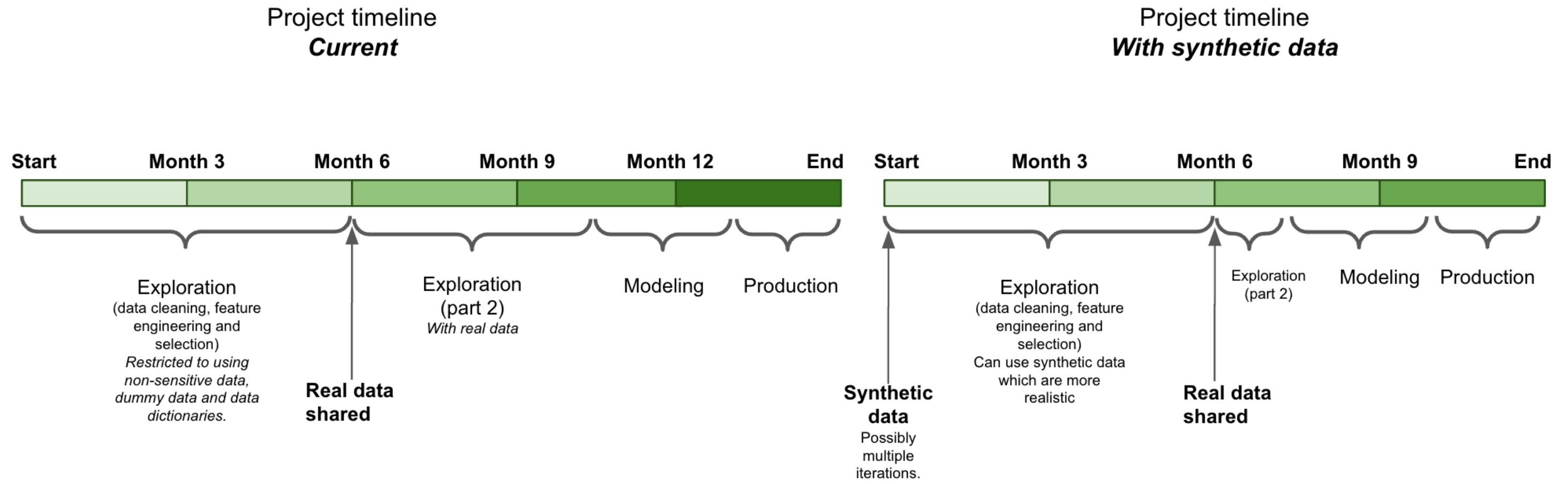}
  \caption{Research projects that use sensitive data often have long waiting times to get access to these data. During this time, initial exploratory analysis can proceed only using publicly available data or dummy data, both of which are typically not good replacements for real data. Synthetic data could potentially be a higher-quality replacement at this exploration stage, allowing for better outcomes and shortening the total project time. This paper explores to what extent synthetic data can replicate the feature importance characteristics of a real dataset, which is one of the key elements in the exploration stage.}
  \label{fig:timeline}
\end{figure*}

The utility of synthetic data inevitably depends on the application for
which this data is intended, and this is reflected in the variety of 
comparative measures of utility.  This may in turn influence the trade-off with privacy that one might be willing to make.  There is no single measure of utility that guarantees the usefulness of synthetic data for all conceivable purposes. 

In this paper, we propose a set of novel measures of utility for synthetic
data to be used in an \emph{exploratory data analysis} setting, specifically focused on
\emph{feature importance}, typically occurring during the early stages of an analysis  (Figure \ref{fig:timeline}). Feature importance describes a number of techniques for scoring or ranking the contribution of each feature in a dataset towards a prediction, when performing predictive modelling.
We assess the privacy--utility trade-off based on these measures, for two differentially-private methods of synthetic data 
generation, \textsc{pate-gan}~
\citep{jordonPATEGANGeneratingSynthetic2019} and PrivBayes
\citep{zhangPrivBayesPrivateData2017},  using three data sets.

In an exploratory setting, the work of data analysis proceeds iteratively and typically involves the following steps: Definition and refinement of the research question, data collection, data cleaning, feature engineering and selection, model building, fitting and evaluation \citep{polyzotis2018data, lam2017button}.  Synthetic data is not always able to be substituted for the original data in all of these steps \citep{stadler2020}, but can it be substituted for \emph{part} of the process?  An affirmative answer to this question would allow for substantial practical benefits: It would mean that certain steps may then be conducted in advance of data access being granted, and a larger group of analysts, who do not require access to the data, can work on them. The outcome of any work carried out on the synthetic data can be refined in subsequent iterations with access to the original dataset.

Where it is applicable, feature \emph{selection} and \emph{engineering} (the process of creating useful features for a predictive model) is likely to be one of the first
tasks performed by a data analyst. We focus on this task and consider a class of utility measures for synthetic data, based on the similarity of feature importance for particular modelling tasks, compared with the original dataset on which it is based. We use measures based on both the ranking of features by their importance, and also their importance scores. These measures have not been used before for the purpose of quantifying utility in synthetic datasets. 

We take into account the privacy trade-off by using differentially private \citep{dp} synthesis methods: All of the methods we consider in this paper offer this strong guarantee \citep{privbayes,rosenblatt2020differentially}. The level of differential privacy is specified in terms of a parameter conventionally called $\epsilon$: the implementation of each method that we use allows us to specify the value of $\epsilon$, or determine it from other input parameters of the method.

The value of $\epsilon$ constitutes a \emph{bound} on privacy. From the point of view of a user
of data synthesis methods, who might have a particular `privacy
budget'---a maximum $\epsilon$ that they consider acceptable---the
best published bound is the strongest privacy claim that they can make.  It is
worth noting this distinction, since we are really comparing the
known privacy bounds of these different methods, rather than the best bounds possible. This might be considered to be a better proxy for the inherent privacy protection offered by the methods. We suggest that such a comparison is the most
useful one, though, at least in this practical sense of relating to the choices that a user can make with what is known about the methods.

While the definition of differential privacy is straightforward, the
best choice of $\epsilon$ for a particular
privacy application, and its interpretation in such practical
settings, is less clear \citep{stadler2020,wagner2018}. While there are other definitions of privacy measures that could provide a more interpretable quantification of privacy in some scenarios, considering these is left as a direction of future work, given the widespread use of differential privacy in academic literature and its adoption by industry, in addition to the simplicity of using $\epsilon$ to compare privacy across different synthetic data methods.

Finally, while synthetic data might be desirable in any setting where there is
data, in this paper we restrict our attention to fully-joined tabular relational data (over a broad set of data types).

Section \ref{sec:benefits} gives an overview of the claimed benefits of synthetic data and the ways in which they can be quantified. Section \ref{sec:fi} introduces some definitions of feature importance and proposes three new measures of utility for synthetic data based on these. Sections \ref{sec:experiments} and \ref{sec:results} describe the experiments we have conducted and discuss the main findings, and Section \ref{sec:conclusions} concludes the paper.

\section{Synthetic Data: Privacy and Utility}
\label{sec:benefits}

For most synthesis techniques, generating a synthetic dataset involves fitting a generative model to the original (sensitive) data, and then sampling from this model to produce synthetic records. 

Synthetic data preserve existing relationships and attributes in the original data only as far as these are preserved by the model.  Properties of the data relevant to a particular use case should therefore be assessed---section~\ref{subsection:utility} reviews some existing techniques.  Similarly, since the generative model has been fit to the original data, information can be leaked by the synthetic records: the next section describes techniques for ensuring that this is to an acceptable degree.

\subsection{Privacy}
\label{subsection:privacy}

A primary motivation for using synthetic data is for its privacy benefit.  One aim of producing synthetic datasets is to be able to grant access to analysts, without having to clear many of the hurdles encountered when working with sensitive data, thus permitting access to a larger group of individuals---potentially the general public.
Several such initiatives have been proposed and implemented in recent years, especially in health research where data are traditionally hard to access \citep{mihaela}\footnote{Also see \url{https://odileeds.org/events/synae/}, \url{https://healthdatainsight.org.uk/project/the-simulacrum/} and \url{https://cprd.com/content/synthetic-data}}.  For these settings, thinking carefully about privacy is of paramount importance.

Defining privacy is not straightforward. A variety of definitions and related measures have been proposed, none of which is universally accepted, or even applicable in all cases.  These fall into three broad categories: 
\begin{enumerate}
    \item Measures that capture the empirical probability of successful identification attacks assuming a motivated intruder with access to certain information \citep{reiter2009estimating, stadler2020, oprisanu2021measuring, logan, monte};
    \item Measures that quantify properties of the released data which are a proxy for privacy---for example, $k$-anonymity \citep{kanonymity,wagner2018,distance_privacy};
    \item Formal guarantees of indistinguishability provided by the generating mechanism, which can be used to place an upper bound on the probability that an intruder is able to infer that a specific individual was part of the original unpublished data---for example, differential privacy \citep{dwork2006,bindschaedler2017plausible,sdl}. 
\end{enumerate}

Measures of the first and second types are based on assumptions about the information available and technical capability of an intruder. These assumptions might not hold in the real world, where an intruder has more knowledge available to them than was anticipated by the data owner. There are several well-known examples of failures of this kind that have resulted in privacy breaches.  One such is described by \citet{netflix}.

Differential privacy \citep{dwork2006}, which belongs to the third class, is among the more widely used definitions. Several data synthesis methods have been proposed that incorporate differential privacy, making comparisons across methods relatively straightforward.

For differential privacy and similar metrics based on theoretical guarantees, the privacy bound can be difficult to interpret. The typical process involves trading off privacy with utility or combining with other privacy metrics which are more interpretable \citep{expose}. \citet{stadler2020} showed that identification attacks can still be carried out successfully on two differentially private synthetic methods, even when setting the privacy parameter to values ostensibly offering a high degree of privacy.

\subsection{Utility}
\label{subsection:utility}

Synthetic data model the joint distribution of the original data, potentially offering higher-accuracy representations compared to methods like aggregation and perturbation. Synthetic data can thus have high utility for users (e.g. analysts, researchers or other practitioners). Like privacy, utility has a range of definitions depending on the task for which the data are needed, but typically it involves comparing the original and synthetic data. There are broadly two types of utility comparisons. The first captures the overall distributional similarity between original and synthetic data and aims at quantifying utility across a range of potential tasks. The second captures similarity when performing specific types of analyses with the data and can be particularly useful when the exact tasks to be performed are known. Finally there are some attempts to combine these two types of metrics \cite{utility_combined}.

A non-exhaustive list of the distributional similarity group includes descriptive statistics (e.g. differences between mean, variance, modes, percentiles, etc of original vs. synthetic data) which offer useful but limited information on distributions and relationships; distributions and frequencies, which involve comparing the marginal, joint or conditional empirical distributions of the original vs. the synthetic data using Kullback-Leibler \citep{kullback1951,airline}, total variation distance \citep{privbayes,bindschaedler2017plausible} and similar metrics; correlations, which show if relationships between all variables are maintained (with alternative measures for categorical variables \citep{cramer}); comparing outliers, missing value patterns and values of constraint variables \citep{synthpop}.

The second group of utility metrics focus on specific analytics or data science tasks such as Machine learning (ML) model comparison, where the same ML model is trained on the original and synthetic datasets independently and used to predict a holdout sample from the original, after which prediction accuracy is measured e.g. see \cite{bindschaedler2017plausible, privbayes, stadler2020}; regression/classification inference, which involves training a linear or logistic regression model on the original and synthetic data and comparing estimated coefficients \citep{synthpop, woo2009global}; adversarial models, which involves training a classification model to distinguish between original and synthetic records and evaluating utility with the remaining parts of the datasets; feature ranking and selection which is used in this paper.

The utility of synthetic data for many of the above tasks is subjective as there is no generic definition of \emph{good utility} across different problems. Synthetic data have been shown to have good utility for some of these tasks in specific scenarios \cite{jordonPATEGANGeneratingSynthetic2019, privbayes, synthpop, airline, covid} but there are also less encouraging results in the literature \citep{stadler2020,ons}. There is a lack of evidence on the usefulness of synthetic data in real complex datasets, where there is a variety of tasks that could potentially be performed by researchers and practitioners and thus multiple definitions of utility.

\section{Feature Importance}
\label{sec:fi}
Among the many ways synthetic data can be used, the focus here is on how well they can act as proxies for the original data when performing data science tasks that typically precede the training of predictive models: in particular, quantifying the importance of features in the dataset for, and engineering new ones.

The motivation behind exploring the feasibility of using synthetic data for these tasks comes from well-known issues when performing data analyses on sensitive data. In many real-world research and industrial projects, analysts and researchers might need to wait for months or years to get access to the original data \citep{data_readiness} due to security restrictions. Even when access is granted, this might be provided only to a limited number of individuals and within strictly controlled environments. At the same time, the data science tasks mentioned above often take up a significant part of a data science project \citep{forbes}, typically more than training the final model. If synthetic data are useful for performing these tasks, analysis can start before full access is granted, saving a lot of time and allowing early insights to be gained. The final optimisation of the features and model can then be performed on the original data when they become available (Figure~\ref{fig:timeline}). 

To measure the utility of synthetic data for the above tasks, we propose using the similarity between different feature importance measures as calculated from original and synthetic versions of the same dataset. We formulate three different similarity measures, which have not been used in this context before. One of these measures is proposed for the first time here. We empirically explore their use in artificially generated and real-world datasets. Ideally, the synthesizers generate datasets in which ranks and feature importance scores of features are preserved. In reality, synthetic data may suggest some features are more important than they are when learned from the original data, and others may be ignored. The proposed measures are designed to capture and reveal these cases.

\subsection{Measuring Feature Importance}
\label{subsec:fi}
A number of measures of feature importance are routinely used by analysts to score (or rank) the features of a dataset by importance when predicting a given target. Some of the most widely used are the following:
\begin{enumerate}
    \item Mean Decrease in Impurity (MDI): A measurement of feature importance can be produced while training a random forest model, by attributing the improvement to the split criterion at each split to the splitting variable. The importance of a feature is measured by accumulating the improvement (where the feature was used in) over all the trees in the forest. The ``Gini importance'' or MDI \citep{breiman2001random} is an example of such a feature importance measure when the splitting criterion is the Gini index. Impurity-based importance is biased towards high cardinality features and can therefore be misleading for datasets that contain numerical features and categorical features with a large number of categories. They are computed using only training data and as a result they do not reflect the ability of a feature to make predictions that generalize to the test set. Nevertheless, they are computationally cheap to evaluate during the training of the model.
    
    \item Permutation feature importance (PFI): This method \citep{breiman2001random} involves randomly shuffling a single feature with its importance measured by the decrease in the model's performance. If a feature is important then permuting it should create a large increase in the model's error. This measure prevents attributing high importance to high cardinality features but is more computationally demanding to calculate.  
    
    \item Shapley Additive explanations (SHAP): SHAP \citep{NIPS2017_8a20a862} is a black-box method of computing feature importance scores for a prediction instance as well as its global importance. Utilising results from game theory \citep{shapley1953value}, SHAP measures features importance by calculating how well a model performs for all possible combinations of features and assigning each feature a portion of the total payout based on its average added value across all possible feature combinations to which it was added. SHAP values can be shown to be unique under three key assumptions: \textit{local accuracy}, \textit{missingness} and \textit{consistency} \citep[Theorem 1]{NIPS2017_8a20a862}. The computation of SHAP values is in general an NP-hard problem and therefore in most cases relies upon approximate solutions. However, for tree-based models, an exact algorithm exists that runs in polynomial rather than exponential time (TreeSHAP of \cite{lundberg2018consistent}).
    
\end{enumerate}

In this work, we measure feature importance using the SHAP method due to its wide adoption, solid theoretical foundation in game theory and the fact that it prevents issues related to feature cardinality.

\subsection{Comparing Feature Importance Scores}
\label{subsec:fi_similarity}

In this section, we highlight three different scores of similarity between two feature-importance lists $S$ and $T$: cosine similarity, rank-biased overlap (RBO) \citep{webber2010similarity}, and correlated corrected RBO. Cosine similarity is a popular and simple measure of similarity that can be used to calculate the similarity between importance scores. However, this approach does not considering the ranking of the feature importances (as computed by the methods detailed in Section~\ref{subsec:fi}). This leads us to consider $\mathrm{RBO}$ \citep{webber2010similarity} which accounts for the aforementioned deficit. Unfortunately, $\mathrm{RBO}$ inherently does not consider that features itself can be highly correlated and will penalise if the feature lists are not exactly the same. As a result, we extend RBO to a novel measure of similarity $\mathrm{RBO}^{Cor}$ to correct for this phenomena. To the authors' knowledge, this paper presents the first application of these methods as proxies for the utility of synthetic data.

\subsubsection{Existing methods}

\paragraph{Cosine similarity} Treating the importance score of the features as a vector in a Euclidean vector space, cosine similarity calculates the cosine of the angle between the two score vectors. It is more robust to small changes in feature importance scores than the rank-based methods (which might undergo large changes when features change rank order). Here, we use the Cosine similarity as implemented in scikit-learn \citep{scikit-learn}.

\paragraph{Rank-biased overlap (RBO)} This is a similarity measure for indefinite rankings proposed by \citet{webber2010similarity}. We denote the element at rank $i$ of the list $S$ by a subscript $S_i$ and a sub-list of $S$ is denoted by a slice $S_{c:d}$ be $\{S_i: c \le i \le d\}$. Given two ranked feature-importance lists be $S$ and $T$, the \textit{agreement} between $S$ and $T$ at depth $d$ is given by $A_{S,T,d} := \frac{|S_{1:d} \cap T_{1:d}|}{d}$, that is, the proportion of $S$ and $T$ that are overlapped at depth $d$. Then, RBO is then defined as:
\begin{equation}\label{eq:RBO}
    \mathrm{RBO}(S,T, k):=\sum_{d=1}^k w_d A_{S,T,d},
\end{equation}
where $w_d = (1-p)p^{d-1}$ for some choice of $p$ with $0 < p < 1$. The parameter $p$ is the geometric drop-off in weight with rank, a smaller value weighting the top ranks relatively more strongly. $\mathrm{RBO}$ tends to be sensitive to features swapping in rank, especially in the top features and when $p$ is small. Two rankings are considered to be similar when a small number of the top-ranked features are the same. In our experiments, we use a Python implementation of $\mathrm{RBO}$\footnote{\url{https://github.com/changyaochen/rbo}}.

One drawback of $\mathrm{RBO}$ is that it does not account for correlation between features. For two highly correlated variables, it is possible that one variable is assigned a certain importance in one list, and in the other list, the same importance is assigned to the other correlated variable. In this case, $\mathrm{RBO}$ will penalise the similarity score, even though the features are similar. 

\subsubsection{A New Method: Correlation-Corrected \textrm{RBO}}

We propose a novel similarity measure to mitigate the limitations of \textrm{RBO} described above by accounting for the correlated features in two ranked feature lists. We define the \textit{corrected} agreement between feature lists $S$ and $T$ at depth $d$ as
\begin{equation}
\label{eq:corrected_agreement}
    A^{Per}_{S,T,d}:=\frac{\max_{T'\in\Pi(T_{1:d})}\sum_{i=1}^d |\mathrm{Corr}(S_i, T'_i)|}{d}=\frac{\max_{P \in \mathcal{P}_d}\sum_{i,j=1}^d |P_{ij}\mathrm{Corr}(S_i, T'_j)|}{d},
\end{equation}
where $\Pi(T_{1:d})$ is all possible permutations of feature list $T_{1:d}$, and $\mathcal{P}_d$ is the set of all permutation matrices. It can be seen that if all features are pairwise uncorrelated then the corrected agreement score is equal to the previous agreement score $A_{S,T,d}$.  In the more likely case where features are correlated, the corrected agreement will not penalise the score as before and are instead treated as a partial overlap.  Similarly to $\mathrm{RBO}$, we define the \textit{permutation-correlation-corrected} RBO as
\begin{equation}
    \mathrm{RBO}^{Per}(S,T,k) := \sum_{d=1}^k w_d A^{Per}_{S,T,d}.
\end{equation}
The corrected agreement (see Equation \ref{eq:corrected_agreement}) is defined in terms of optimising  over all possible permutations of the feature list $T_{1:d}$ which involves a search over $d!$ elements. This quickly becomes prohibitively expensive for large $d$. Instead, we consider optimisation over a large class of matrices $\tilde{\mathcal{P}}$ that is the set of matrices that are doubly stochastic (i.e., it's rows and columns sum to one). More precisely, we have the \textit{correlation-corrected agreement}
\begin{equation}
\label{eq:corrected_agreementq}
    A^{Cor}_{S,T,d}:=\frac{\max_{\tilde{P} \in \tilde{\mathcal{P
    }}}\sum_{i=1,j=1}^d | \tilde{P}_{ij} \mathrm{Corr}(S_i, T_j) | }{d}.
\end{equation}
From Birkhoff–von Neumann theorem, doubly stochastic matrices can be decomposed into linear combination of permutation matrices whose coefficents each is greater than zero and all sum to one. Thus, doubly stochastic matrices can be interpreted as soft assignment over permutation matrices and does not force a one-to-one correspondence between features. A computational advantage of considering $\tilde{\mathcal{P}}$ is that we can calculate the corrected agreement using linear programming. In our implementation, we use the PuLP solver\footnote{\url{https://github.com/coin-or/pulp}}.
Note that all permutation matrices are doubly stochastic and so we have the relationship $A^{Cor}_{S,T,d}\ge A^{Per}_{S,T,d} \ge  A_{S,T,d}$ between the each corrected agreement with equality when all the features are independent of each other. Similarly to before, we define \textit{correlation-corrected} RBO is defined as
\begin{equation}
    \mathrm{RBO}^{Cor}(S,T,k) := \sum_{d=1}^k w_d A^{Cor}_{S,T,d}.
\end{equation}

\section{Methods and Datasets}
\label{sec:experiments}

In this section, we describe the series of experiments performed. We use two popular synthetic data generation methods to generate synthetic data based on five artificial and three real datasets. We then apply the proposed feature importance measures and similarity measures of Section \ref{sec:fi} in order to:

\begin{itemize}
    \item Assess whether synthetic data can be useful (produce similar results to real data) when doing feature engineering and selecting the most important features;
    \item Compare the proposed feature importance measures and similarity measures;
    \item Demonstrate how synthetic data generation methods differ and how feature importance preservation varies when changing the privacy budget. 
\end{itemize}

\subsection{Synthesis Methods}
Both of the synthetic data generation methods we use estimate the parameters of the method from the original dataset in a differentially private way. This means that the resulting data synthesisers can be used to generate multiple differentially private synthetic datasets with no additional loss of privacy.

\subsubsection{PrivBayes}
PrivBayes \cite{zhang_privbayes_2014,zhangPrivBayesPrivateData2017} is a differentially private method to approximate the joint probability distribution of a sensitive data set using a set of conditional probabilities between lower dimensional subsets of features from the original data set. The method constructs a Bayesian network connecting each feature of the original data set to up to \emph{k} parent features and estimates the conditional probability distribution for each feature and its restricted set of parent features. Both the determination of the network structure and the estimation of the conditional probability distributions are done in a differentially private way. Synthetic data sets can then be generated by sampling from the differentially private approximate distribution represented by this Bayesian network.

In addition to the differential privacy budget $\epsilon$, PrivBayes has two key parameters. The first is \emph{k}, the maximum number of parents each feature is permitted in the structure of the network. The second is $\beta$, the proportion of the differential privacy budget that is spent on determining a good network structure versus estimating the conditional probability distributions. In this work we use the PrivBayes implementation from \cite{ping_datasynthesizer_2017}.

\subsubsection{PATE-CTGAN}
\cite{rosenblatt2020differentially} applied the PATE (Private Aggregation of Teacher Ensembles) framework of \cite{papernot2017semisupervised} to CTGAN \citep{xu2019modeling}, a state-of-the-art GAN model for synthesizing tabular data. We use the PATE-CTGAN implementation of \textit{smartnoise-sdk}\footnote{https://github.com/opendifferentialprivacy/smartnoise-sdk} and \textit{privgem}\footnote{Refer to \url{https://github.com/kasra-hosseini/privgem} \citep{privgem_2022}.}. 

In PATE-CTGAN, a dataset is partitioned into \emph{k} subsets, and \emph{k} differentially private teacher discriminators are trained. In all our experiments, we set the number of samples per teacher to 1000. Therefore, the number of teachers and data partitions, \emph{k}, change depending on the size of dataset (see Section~\ref{subsec:datasets} for details on the datasets). We set the teacher and student iterations to five ($n_{T} = n_{S} = 5$ as in \cite{jordonPATEGANGeneratingSynthetic2019}). The models are trained using the Adam Optimizer with the learning rate of $2 \times 10^{-4}$ and batch size of 64 (for both the generator and discriminator). 

We trained 25 PATE-CTGAN models for each value of the privacy budget $\epsilon$. Each model was trained for approximately 3000 iterations regardless of the privacy budget. This was achieved by varying two hyperparameters: \textit{noise multiplier} (defined as the inverse exponential decay parameter in the Laplace distribution, $1 / \lambda$) and \textit{moments order} (the order of moments to keep in \textit{moments accountant} of \citet{Abadi_2016}). Table~\ref{table:pate-ctgan-params} summarises these parameters for each value of $\epsilon$. 

\begin{table}[]
\centering
\caption{Performance of GAN models vary significantly as a function of training iterations. In our experiments, we set two hyperparameters: \textit{noise multiplier} and \textit{moments order} such that the number of training iterations is comparable across models with different privacy budget $\epsilon$.}
\begin{tabular}{|c|l|c|c|c|}
\hline
$\epsilon$ & $\delta$  & noise multiplier       & moments order & training iters \\ \hline
0.001      & $10^{-5}$ & $1.05 \times 10^{-7}$  & 30000         & 3077           \\
0.01       & $10^{-5}$ & $1.1 \times 10^{-6}$   & 10000         & 2802           \\
0.1        & $10^{-5}$ & $1.1 \times 10^{-5}$   & 1000          & 2791           \\
0.4        & $10^{-5}$ & $4.2 \times 10^{-5}$   & 1000          & 3025           \\
1          & $10^{-5}$ & $1.05 \times 10^{-4}$  & 100           & 2949           \\
4          & $10^{-5}$ & $3.9 \times 10^{-4}$   & 100           & 3054           \\
10         & $10^{-5}$ & $9.0 \times 10^{-4}$   & 100           & 2971           \\ \hline
\end{tabular}
\label{table:pate-ctgan-params}
\end{table}

\subsection{Datasets}
\label{subsec:datasets}
We use five artificially generated and three real-world datasets to evaluate the usefulness of various synthetic methods for capturing feature importance.  The particular real-world datasets were chosen because they offer a range of different attributes (personal, household, longitudinal), are publicly available, readily accessible, and have been utilised widely by the data science community so are well documented.

\subsubsection{Adult}
The Adult dataset\footnote{UCI Machine Learning Repository \cite{Dua:2019}: http://archive.ics.uci.edu/ml} was extracted from the 1994 US Census Bureau database and has been widely used in the ML community. The prediction task is to determine whether a given adult makes more than \$50,000 a year based on 13 attributes such as education, marital status, occupation, sex, hours of work per week, etc. This dataset has 32,561 rows and 14 columns.

\subsubsection{Household Poverty}
The Costa Rican Household Poverty dataset\footnote{https://www.kaggle.com/c/costa-rican-household-poverty-prediction} contains characteristics for a sample of 9,557 individuals from Costa Rican households. The features include observable household attributes like the material of their walls and ceiling, assets found in the home, educational attainment, demographic data, etc. We removed several features of low importance from the dataset and converted the one-hot-encoded features to label-encoded, resulting in a dataset with 29 columns. The prediction task is to determine the poverty level of each household, which is a four-class classification problem. It is possible to combine/aggregate the characteristics of individuals in each household to engineer new features that increase predictive accuracy. In contrast to the other two datasets, this one contains a hierarchical structure (households-individuals). The synthetic algorithms we use do not capture this hierarchy as will be shown later.

\subsubsection{Polish survey}
This Polish survey dataset\footnote{http://www.diagnoza.com/index-en.html} contains answers to a long-running quality of life survey conducted in Poland, under the Social Diagnosis project. The data are collected every two years from specific households in order to monitor how their attitudes, mind-sets and behaviours change over time. We make use of a version of the 2011 dataset which contains 4,895 individuals of age 16 and over and 13 variables for each individual including sex, age, education, marital status, income, etc. We extracted this version of the 2011 dataset from the synthpop R library found in CRAN \cite{synthpop}. The prediction task is to determine the intention of an individual to work abroad given all other characteristics, which is a binary classification task.


\section{Experiments}
\label{sec:results}

In this section, we describe the results of the experiments detailed in the previous section.  We first discuss results from the artificially generated datasets, which are designed to reveal how different characteristics in a dataset impact the ability of synthetic data to effectively replicate the importance of each feature found in the original data.  We then consider the results on the three real-world datasets.

\begin{landscape}
\begin{figure}%
    \centering
    \subfloat[\centering 1]{\includegraphics[width=0.31\columnwidth]{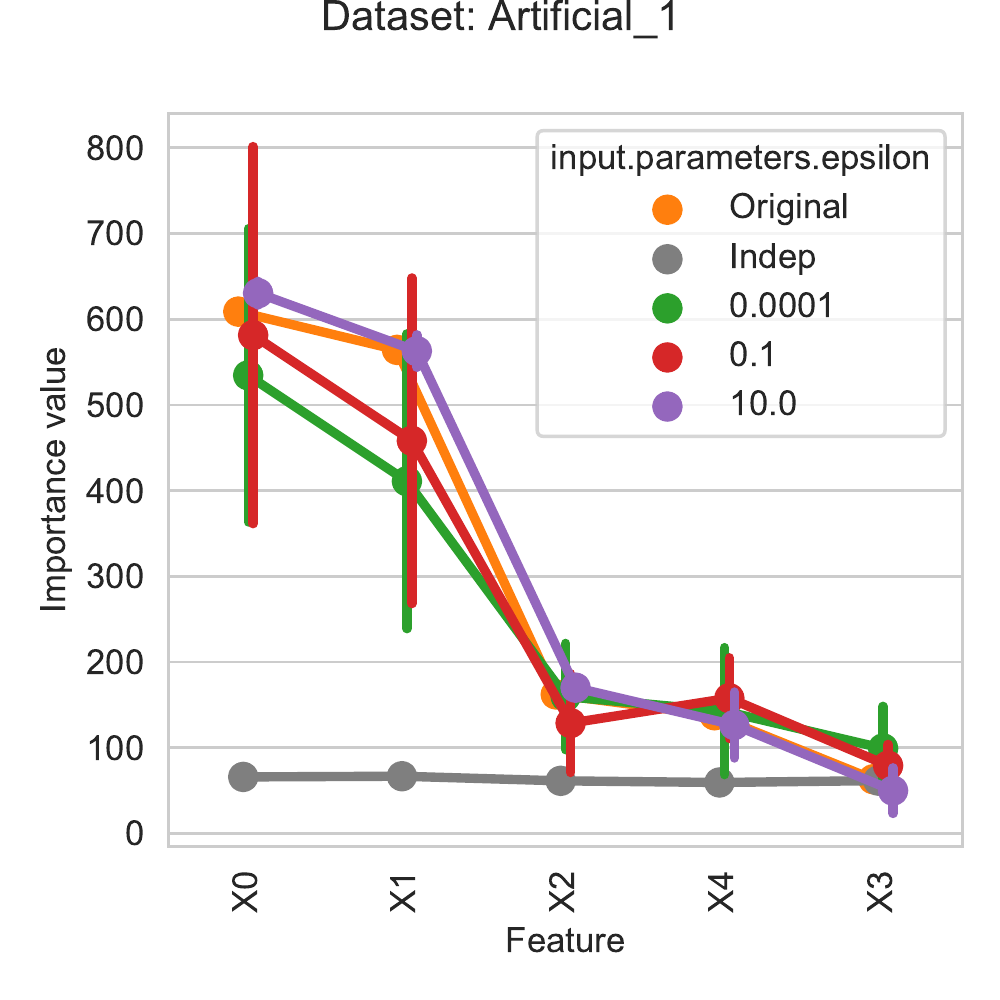}}%
    \subfloat[\centering 2]{\includegraphics[width=0.31\columnwidth]{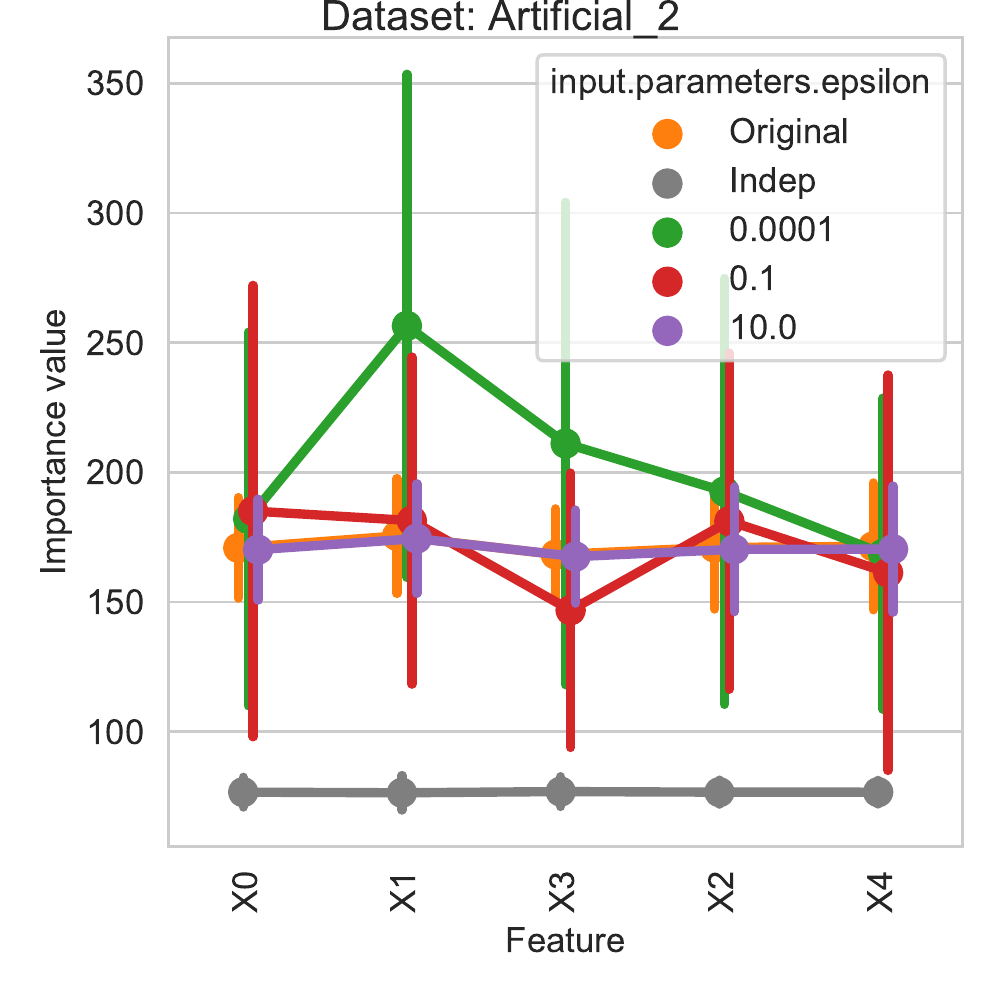}}%
    \subfloat[\centering 3]{\includegraphics[width=0.31\columnwidth]{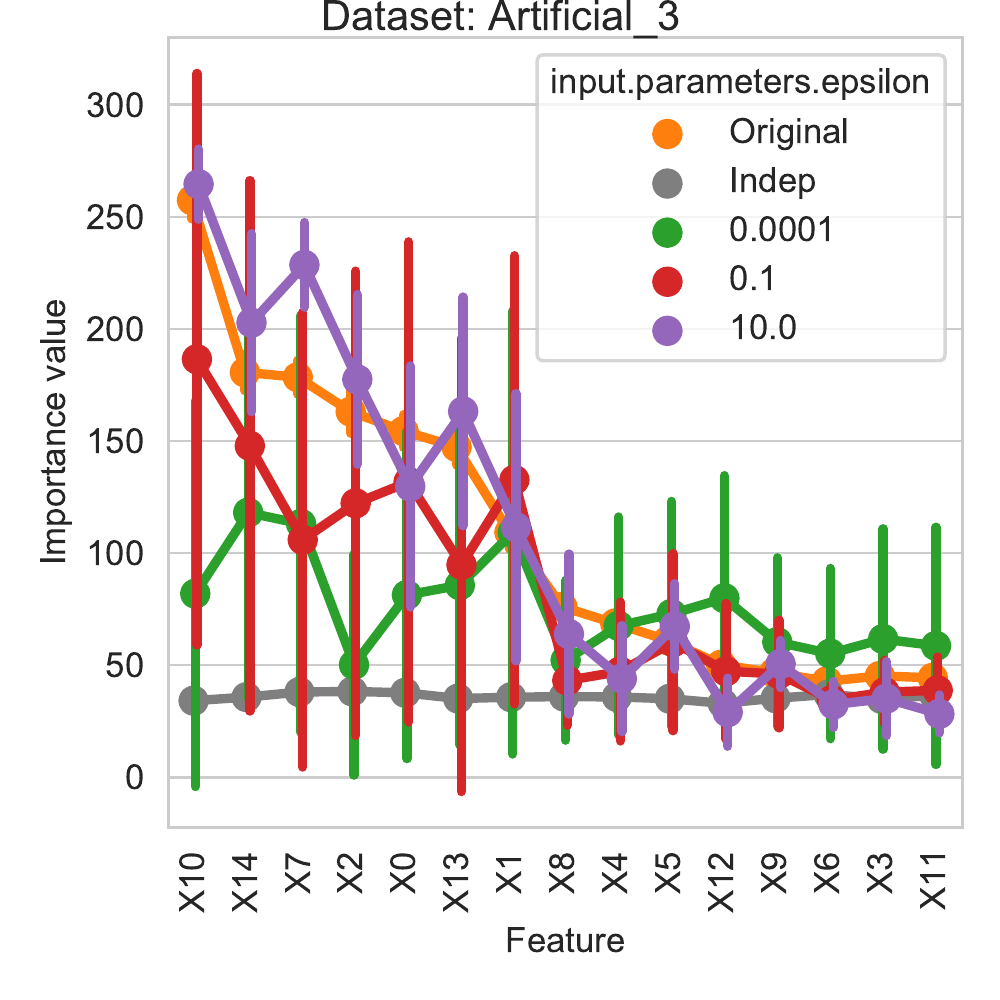}}%
    \qquad
    \subfloat[\centering 4]{\includegraphics[width=0.31\columnwidth]{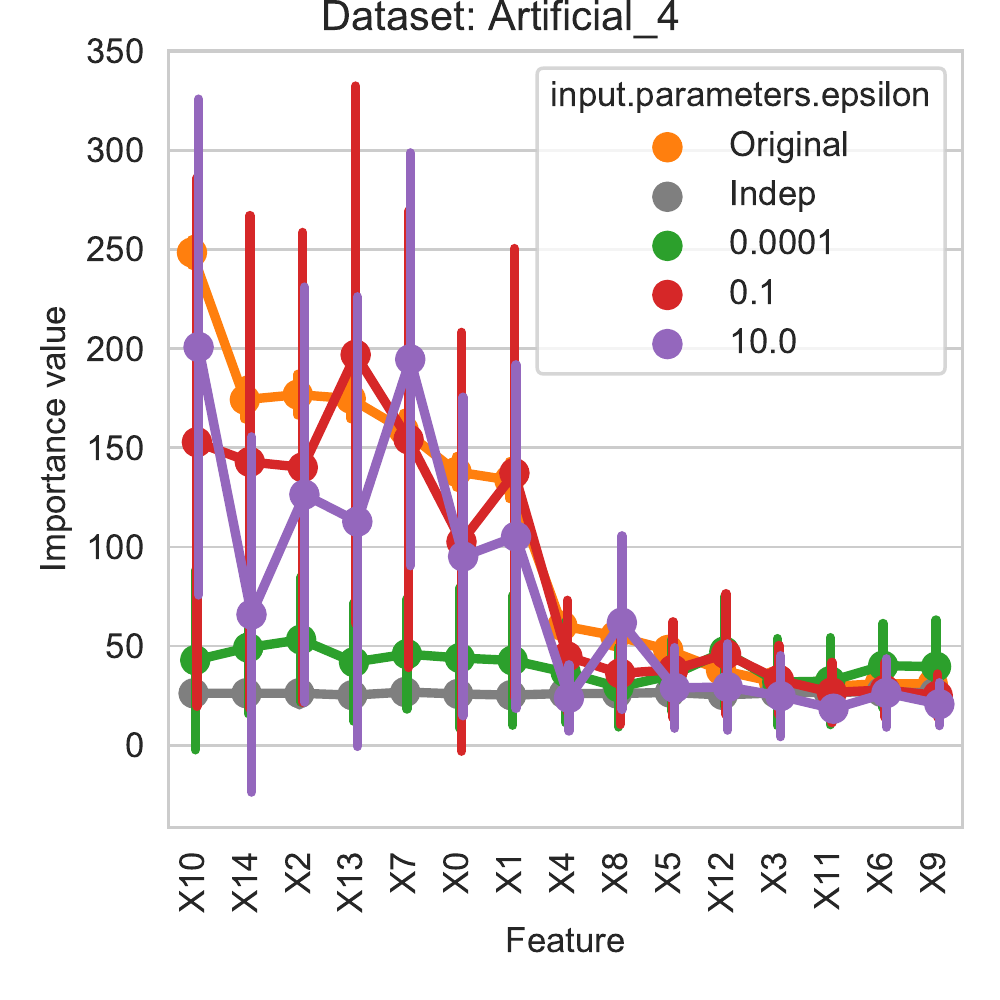}}%
    \subfloat[\centering 5]{\includegraphics[width=0.31\columnwidth]{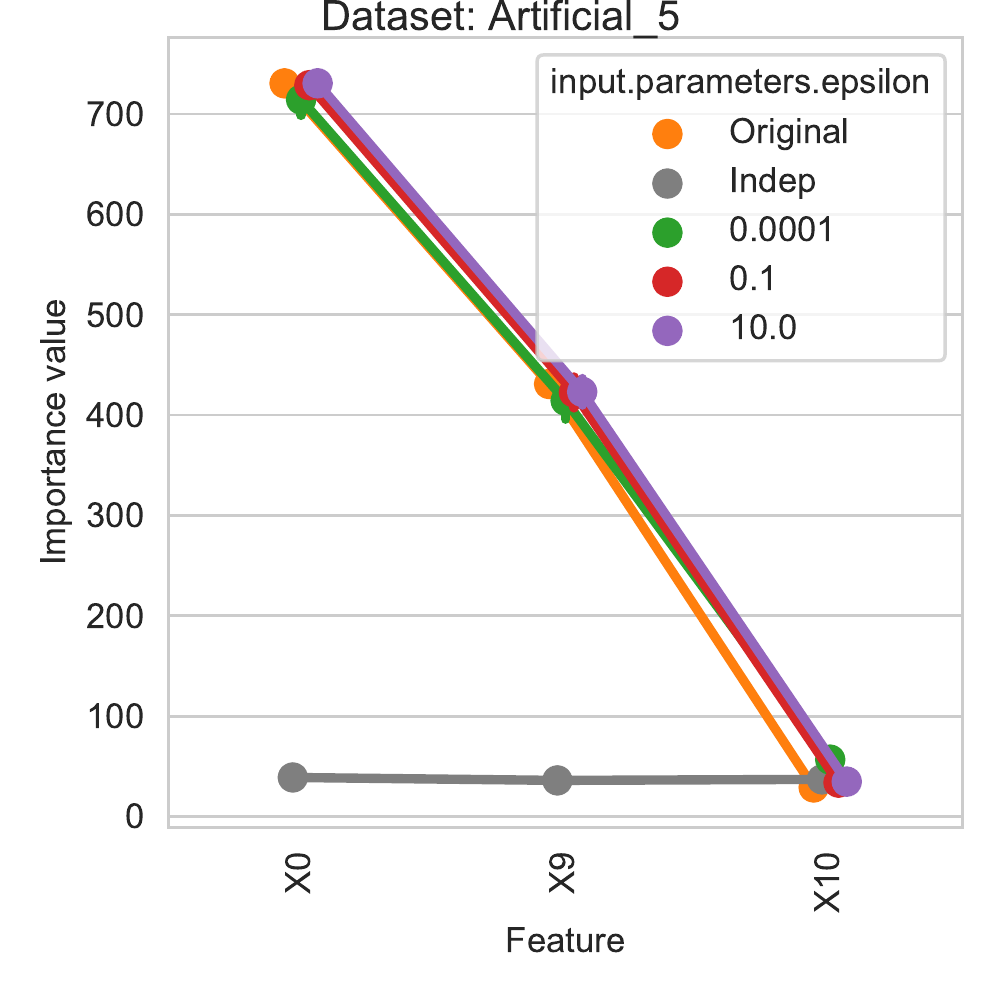}}%
    \caption{Artificial datasets 1--5: Feature importance (SHAP) values from the original dataset, the PrivBayes synthetic datasets (for three  values of the privacy parameter $\epsilon$) and the baseline synthetic dataset computed by independent column resampling. The lines show the mean and standard deviation from 25 independent runs. It is much easier for feature importances to be ranked correctly using synthetic data when ``real'' importance values are clearly separated (as e.g. in Artificial 1 and 5). This becomes harder in Artificial 2 and in datasets with large numbers of features where there is overlap in feature importances (Artificial 3 and 4).}%
    \label{fig:fi_feature_importance_artificial_data}%
\end{figure}
\end{landscape}
\subsection{Artificial Data}

We generate five artificial datasets with characteristics that allow us to test the similarity performance for a range of scenarios. All the datasets were created using scikit-learn's \citep{scikit-learn} \texttt{make\_classification} method and have 10000 rows. They are all designed to be used for binary classification. In more detail:
\begin{description}
    \item[Artificial 1] This dataset has five informative categorical features (features that convey useful information for predicting the output variable). Each feature can take on one of five categorical values. This is used as the baseline case.
    \item[Artificial 2] This dataset has five categorical features with five categories each. Any one feature is informative by itself, and the remaining four are redundant. This dataset is used to demonstrate what happens when all features have virtually the same importance (which happens in cases of strong multi-collinearity).
    \item[Artificial 3] This dataset has 15 categorical features with five categories each. All features are informative. This dataset is used to demonstrate how feature importance similarity scores are affected when the number of features increases. 
    \item[Artificial 4] This dataset has 15 continuous numerical features. All features are informative. This dataset is used to demonstrate how feature importance similarity scores are affected when features are continuous numerical rather than categorical.
    \item[Artificial 5] This dataset has three categorical features with 5 categories each. The features were designed to clearly differ in importance for the prediction task. This will help demonstrate how similarity scores change when the importance of two features become more separated---that is, when the feature ranking task becomes easier.
\end{description}


A first set of comparisons can be made by considering Figure~\ref{fig:fi_feature_importance_artificial_data}. These show the Shapley (or SHAP) values, our chosen measure of feature importance, for each feature of Artificial datasets 1, 2 and 5.

\textbf{Artificial 1} simulates a typical scenario where some features are more important for the classification task than others; there are two features ($X_0$ and $X_{1}$) with high importance and three with lower importance, with very small variance in the importance scores between independent runs (the variance originates from the random split between training and validation samples and from the Random Forest training process). The means of the PrivBayes importance score are close to the scores from the original dataset, but with clearly larger variance between runs, and variance decreasing with $\epsilon$, as expected. The synthetic datasets (for all choices of $\epsilon$ we considered) allow us to correctly infer that features $X_0$ and $X_{1}$ are more important than features $X_{2}$, $X_{3}$ and $X_{4}$, but not their ranking within these groups, other than for $\epsilon = 10.0$, the largest value we considered. Independent column resampling (grey line) gives similar importance to each feature, which is expected: this baseline model does not capture any relationships between the features and the output variable, and all features consequently have similar, low importance for predicting the output variable.

\textbf{Artificial 2} simulates an extreme scenario, where each feature is a linear combination of the others. As seen in Figure \ref{fig:fi_feature_importance_artificial_data}, the original data (orange line) have feature importance values that are very similar, with variance between runs that is larger than the variance observed with Artificial 1.  The mean importance scores of PrivBayes follow a similar trend to Artificial 1, with the exception of $\epsilon=0.0001$, the smallest value we consider. where the synthetic version is so distorted that feature $X_{1}$ appears to be more important than the others. In this artificial experiment, it is clear that it is very difficult to capture a stable ranking of features that appears in the original dataset, and this is entirely determined by random variation between runs, so comparing feature imporance purely by their rank will be misleading.  The independent column resampling baseline is expected to perform better in this task than the Artificial 1 task, as it always leads to close-to-uniform feature importance scores.

\textbf{Artificial 5} simulates an `easy' situation for replicating feature importance: There are three features ($X_0$, $X_{9}$, $X_{10}$) constructed to have distinct and well-separated feature importance values. As can be seen in Figure \ref{fig:fi_feature_importance_artificial_data}, all PrivBayes synthetic datasets lead to feature importance scores very similar to the original data, and variance is much lower than in the Artificial 1 and 2 tasks. In this case, the ranking of features is reproduced correctly for each epsilon bound, and we expect to see high similarity scores for both rank-based and score-based similarity measures. Independent column resampling again leads to uniform importance scores and is expected to score poorly in similarity with the original dataset.

\begin{landscape}
\begin{figure}[ht]
  \centering
 \includegraphics[width=1.0\columnwidth]{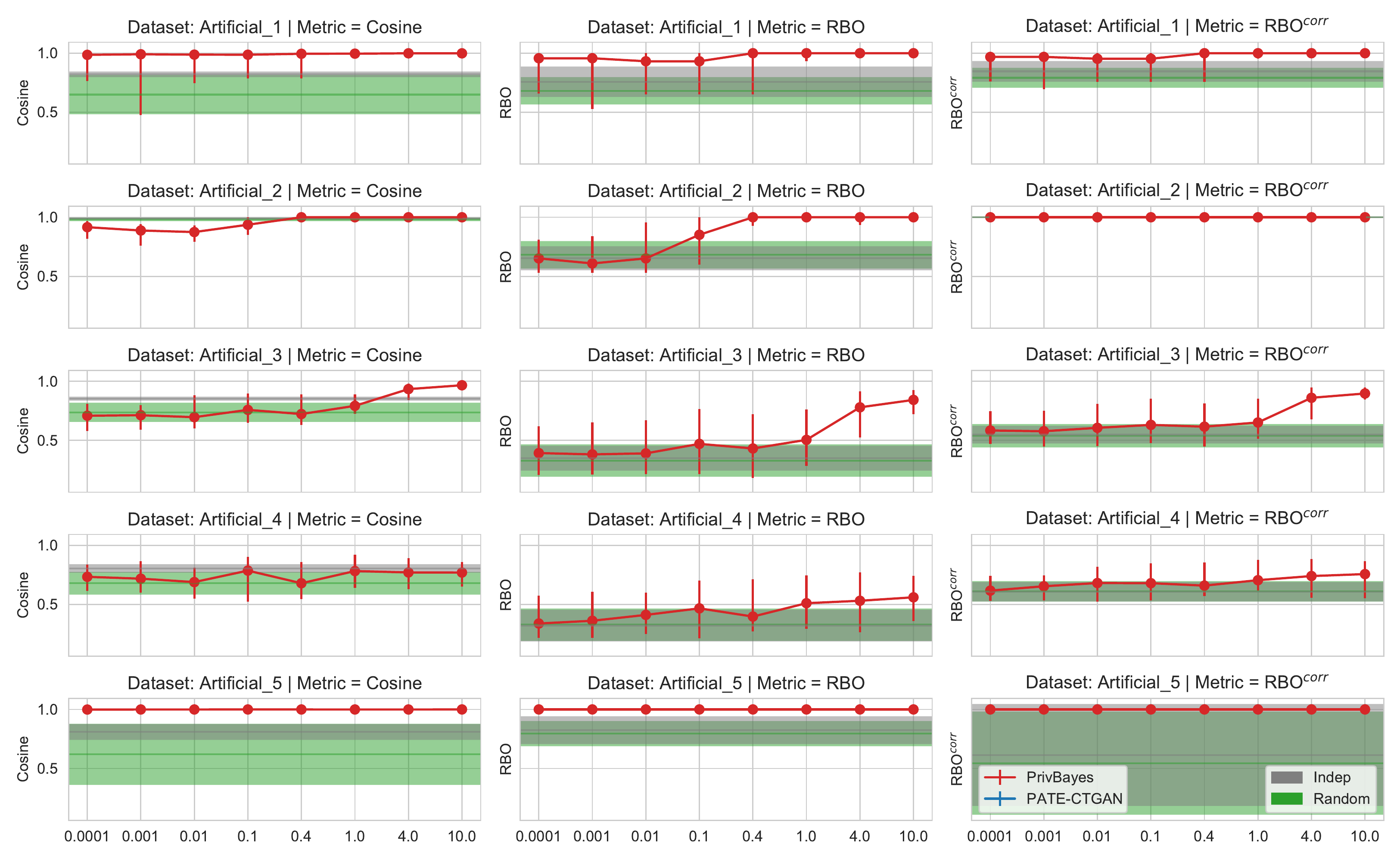}
    \caption{Artificial datasets 1--5: Feature importance similarity measures (Cosine,  $\text{RBO}$, $\text{RBO}^{\text{Corr}}$) vs. DP $\epsilon$.  These measures computed on the original datasets are compared with PrivBayes synthetic datasets (horizontal axis), the independent column resampling baseline (grey ribbon), and the random feature permutation baseline (green ribbon). The lines show the mean and $\pm$1 standard deviation computed from 25 independent runs. Synthatic data derived from datasets with clear feature separation and a smaller number of features have better feature importance similarity scores.}
    \label{fig:fs_artificial}
\end{figure}
\end{landscape}

As has been discussed so far, Figure \ref{fig:fi_feature_importance_artificial_data} shows the feature importance (SHAP) scores, and how well these can be captured by synthetic data. To summarize the importance scores for each feature into a single number, we employ the measures of similarity described in Section \ref{subsec:fi_similarity}.  The results for the same tasks on our artificial datasets are shown in Figure \ref{fig:fs_artificial}. The three columns in the figure show feature importance similarity between the original and the synthetic data, as measured by $\text{RBO}$, $\text{RBO}^{\text{Corr}}$ and Cosine similarity, for each artificial dataset, using the PrivBayes synthesis algorithm. Each subplot shows how similarity varies with $\epsilon$. Each point represents the mean of 25 independent runs, with the standard deviation also shown. The shaded areas represent the means and standard deviations of the two baselines described in the previous section (independent column resampling and random feature permutation).

From this figure, it is apparent that for the classification task on  the Artificial 1 dataset, synthetic data generated by PrivBayes achieves a high similarity score by all measures, and with very small variance, when taking $\epsilon > 0.4$.  For smaller $\epsilon$, the variance increases and performance starts to degrade towards the two baseline methods, which is especially apparent for the two rank-based similarity measures ($\text{RBO}$ and $\text{RBO}^{\text{Corr}}$). The lower similarity and larger variance can be understood with reference to Figure~\ref{fig:fi_feature_importance_artificial_data}, comparing the overall uncertainty in importance scores from the synthetic data.

Comparing Artificial 1 and Artificial 2, it is clear that the RBO measure of feature importance similarity is degraded (particularly for smaller values of $\epsilon$).  This is to be expected, as explained previously: this dataset has several correlated features, so their ranking is unstable.  Cosine does not depend explicitly on the ranking of the feature importance scores, and $\text{RBO}^{\text{Corr}}$ corrects for the correlation, so there are more reliable indicators of similarity.

Finally, the similarity scores of Artificial 5 are all very close to 1, and all have very small variance. The importance scores differ substantially across the features, and the synthetic data does a good job of replicating them, even for very small values of $\epsilon$.

Artificial datasets 3 and 4 have been designed to demonstrate how similarity scores are affected when the number of features increases. These have 15 categorical (3) or continuous (4) features, compared with the five of Artificial dataset 1.

The Artificial 3 subplot of
Figure~\ref{fig:fi_feature_importance_artificial_data}, showing the
feature importance values for every feature separately, and
demonstrates that the original feature importances vary, with one
feature being clearly the most important, followed by two clusters of
features having `medium' and `low' importance. The PrivBayes synthetic
dataset version preserves these groupings well when using $\epsilon =
10.0$, and to a certain extent when using $\epsilon = 0.1$. The
variance across runs is large in all cases, which is expected to lead
to many rank swaps between independent runs.  A similar picture
emerges from the Artificial 4 subplot of the same figure.

Owing to the above, there is a visible deterioration across similarity measures for Artificial datasets 3 and 4 compared to Artificial dataset 1. Figure \ref{fig:fs_artificial} shows that all measures drop drastically and significantly overlap with the baselines. Artificial 4 performs slightly worse than Artificial 3, especially for large values of $\epsilon$.


\subsection{Real-World Data}

\begin{landscape}
\begin{figure}%
    \centering
    \subfloat[\centering Adult]{{\includegraphics[width=0.33\columnwidth]{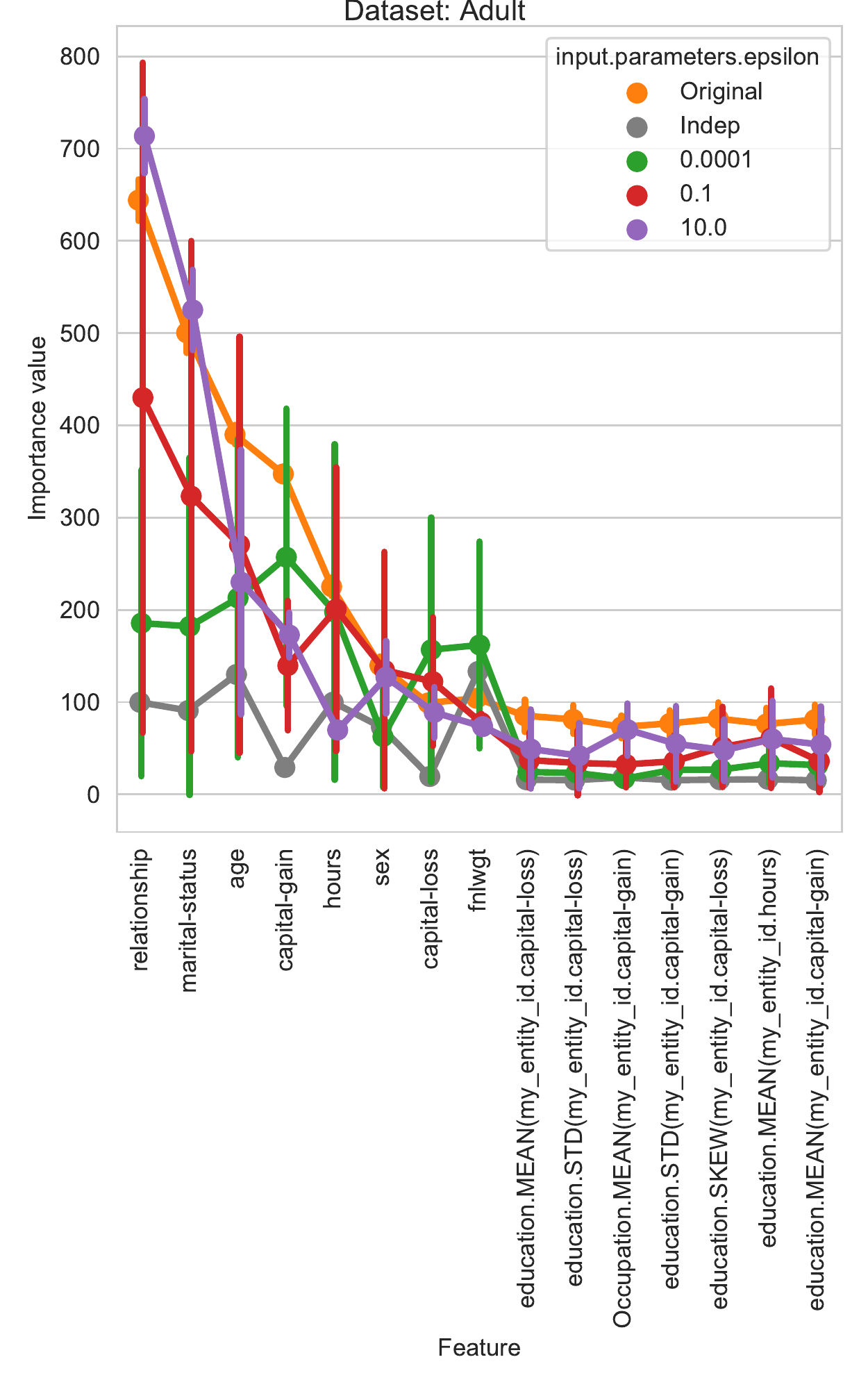}}}%
    \subfloat[\centering Household]{\includegraphics[width=0.33\columnwidth]{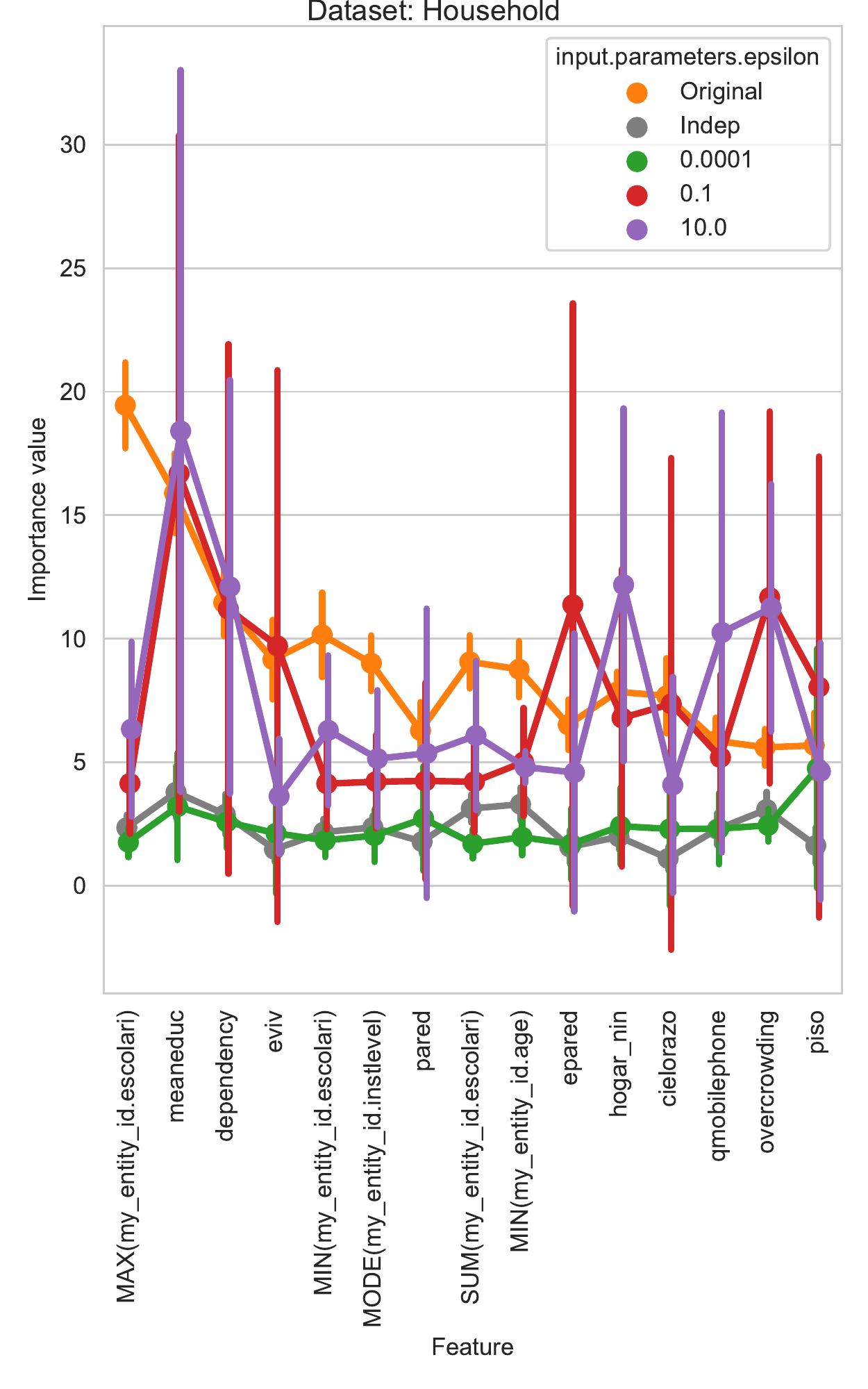}}%
    \subfloat[\centering Polish]{\includegraphics[width=0.33\columnwidth]{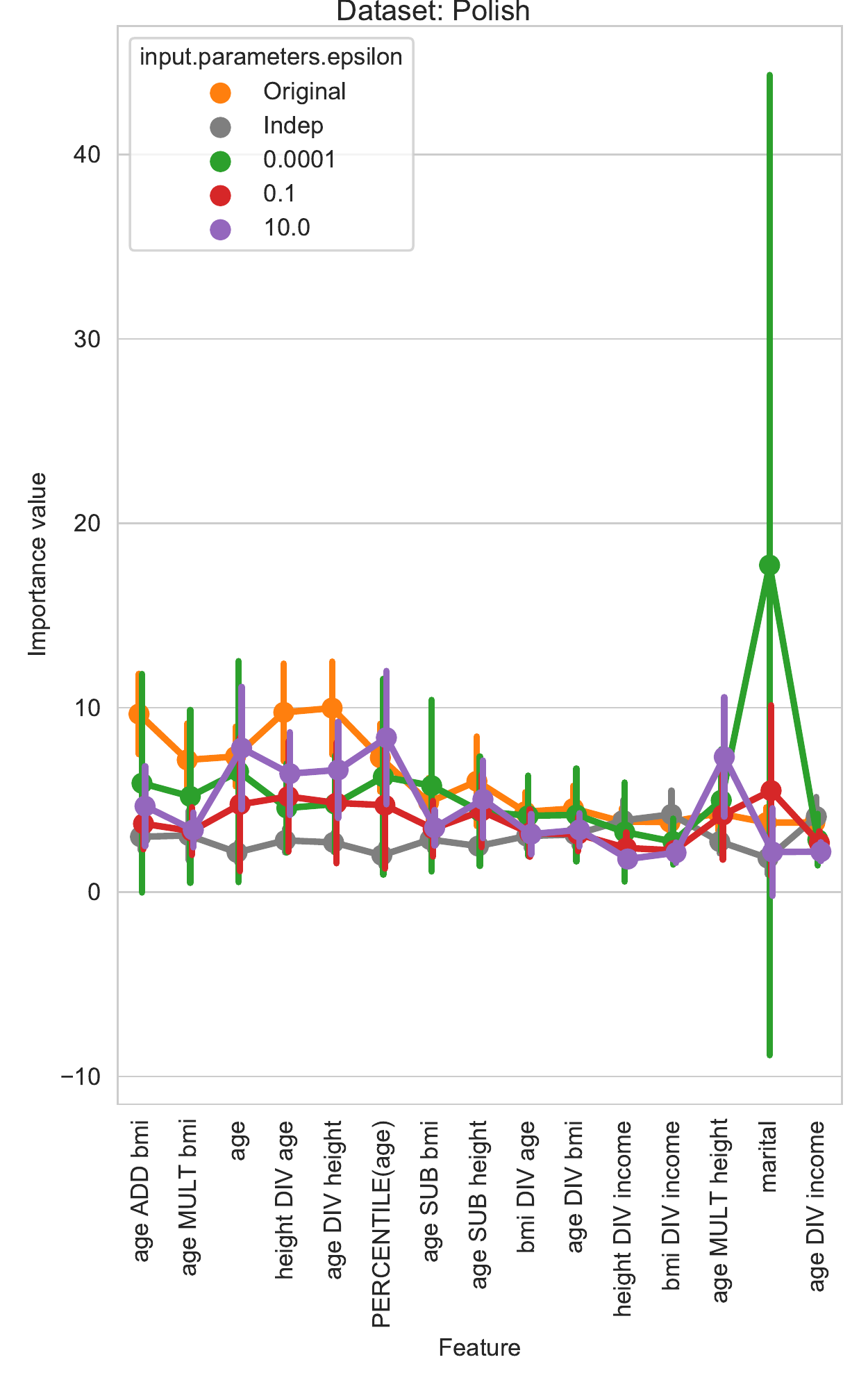}}%
    \caption{Feature importance values from original dataset, PrivBayes synthetic datasets (three different epsilon values) and independent column resampling dataset. The lines show the mean and +/-1 S.D from 25 independent runs.}%
    \label{fig:fi_feature_importance}%
\end{figure}
\end{landscape}

Switching to the three real-world datasets, we first demonstrate which features are the most important for each dataset and how their feature importance scores vary across independent runs. Figure~\ref{fig:fi_feature_importance} shows the mean and standard deviation of the SHAP feature importance scores of the top 15 features for each of the three datasets, with features in order of mean importance when using the original data. Values for the original, baseline and PrivBayes versions are shown, with PATE-CTGAN excluded to simplify the visualisation.

In the Adult dataset (Figure \ref{fig:fi_feature_importance}), the most important features for predicting income with the original data are non-engineered features, especially relationship, marital status, age and capital gain. The engineered features are all less important than the non-engineered ones. The variance in the original data is small but increases a lot when using PrivBayes, especially when epsilon drops. The variance originates mainly from the synthetic data generation process which distorts the dataset's joint distribution, with larger distortion when introducing more noise, as expected. Feature importances in the original data are somewhat distinct, with a steep curve in importance from left to right, although there is some clustering of similar importances after the eighth feature. The synthetic data mean importances partially capture this pattern but seem to fail to do so when epsilon is 0.0001. Independent column resampling is mostly uniform.

In the Household dataset (Figure \ref{fig:fi_feature_importance}) some engineered features are important due to the hierarchical structure of the dataset; it seems there is useful information in summarising data from all individuals in a household when predicting household poverty. The years of schooling that members of the household have gone through seems to be a strong predictor but there is also some variance here.   The separation between feature importances of the features is not very large compared to the Adult dataset. There are 2-3 features with high importance but the remaining ones differ slightly, with some variance even in the original feature importances. The PrivBayes versions seem to completely miss the high importance of engineered features, giving them low importance and with large variance for all epsilon bounds. This is a result of the inability of PrivBayes to capture hierarchical structures in datasets; using PrivBayes effectively leads to a synthetic dataset where households have one member each. It is clear that in this case we expect a lot of rank swaps between features when using the synthetic data, even for the top features which affect rank-based similarity drastically. The Adult dataset is somewhat more stable and swaps do not happen for the top features.

In the Polish dataset (Figure \ref{fig:fi_feature_importance}), both engineered and non-engineered features score high in terms of importance. Importance scores for the original data are close to each other to the point where the exact rank is not fixed (e.g. age can be more or less important than the engineered feature ``age MULT bmi'' depending on the random seed). Synthetic versions seems to partially capture some of the most important features on the left side when epsilon is high and variance of importance values is lower than with the other two datasets. But the overall picture is of a dataset with relatively uniform importances which are challenging to replicate. This is partially explained by the fact that the Polish dataset has the highest multi-collinearity compared to the other two datasets. The average correlation (or correlation-like association for categorical values) between its features is 0.325 compared to 0.149 for the Adult and 0.219 for the Household dataset.

\begin{landscape}
\begin{figure}[ht]
  \centering
 \includegraphics[width=0.98\columnwidth]{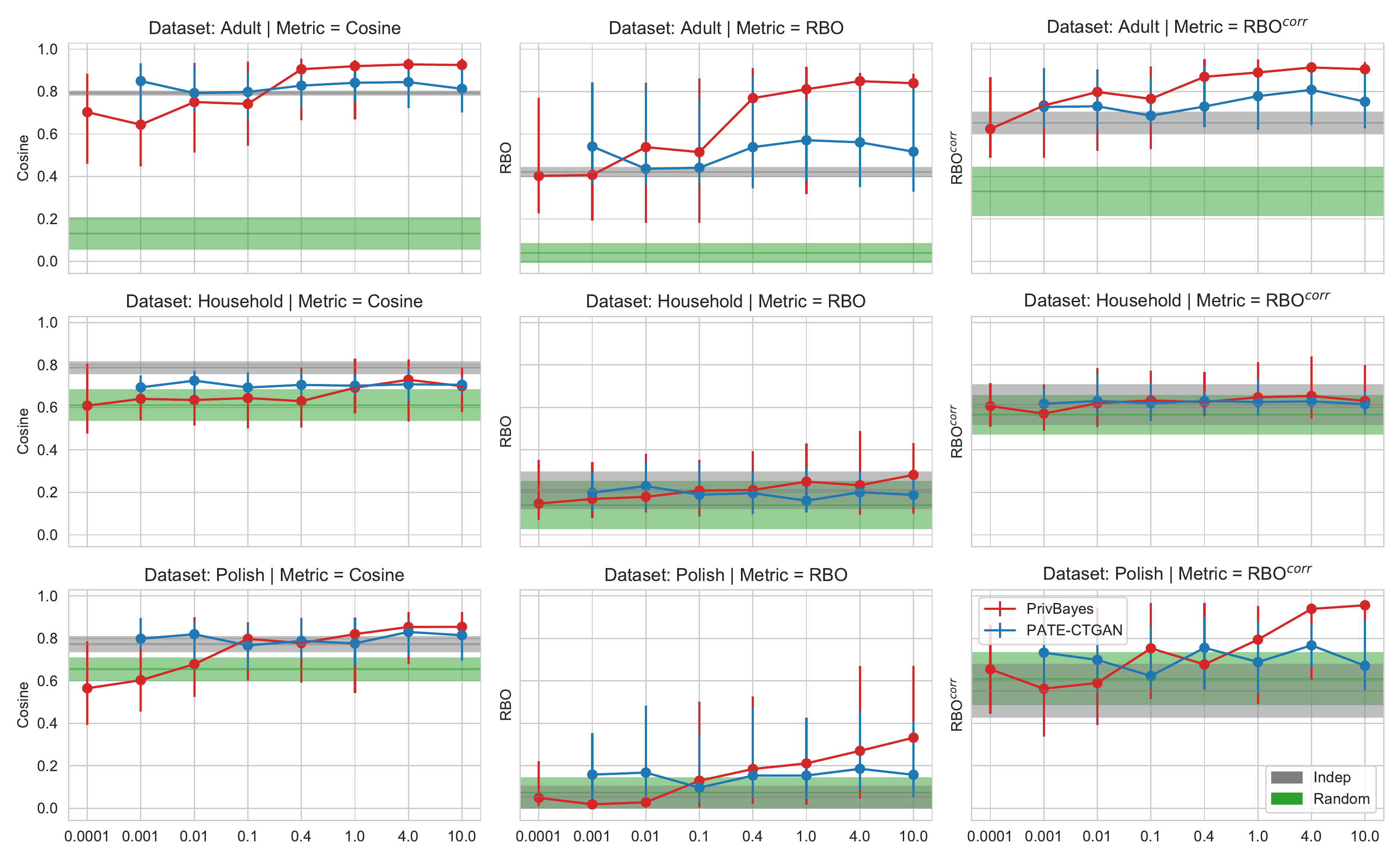}
    \caption{All real-world datasets: Feature importance similarity measures (Cosine similarity,  $\text{RBO}$, $\text{RBO}^{\text{Corr}}$) vs. differential privacy epsilon. Similarity is measured between the original datasets and 1) PrivBayes and PATE-CTGAN synthetic datasets (for all epsilon values). 2) Independent column resampling datasets. 3) Random feature permutation datasets. The lines show the mean and +/-1 S.D from 25-50 independent runs. The Adult dataset has more clear separation at least among its top features.}
    \label{fig:fs_real}
\end{figure}
\end{landscape}

\begin{figure}%
    \centering
    \subfloat[\centering Adult]{{\includegraphics[width=0.35\columnwidth]{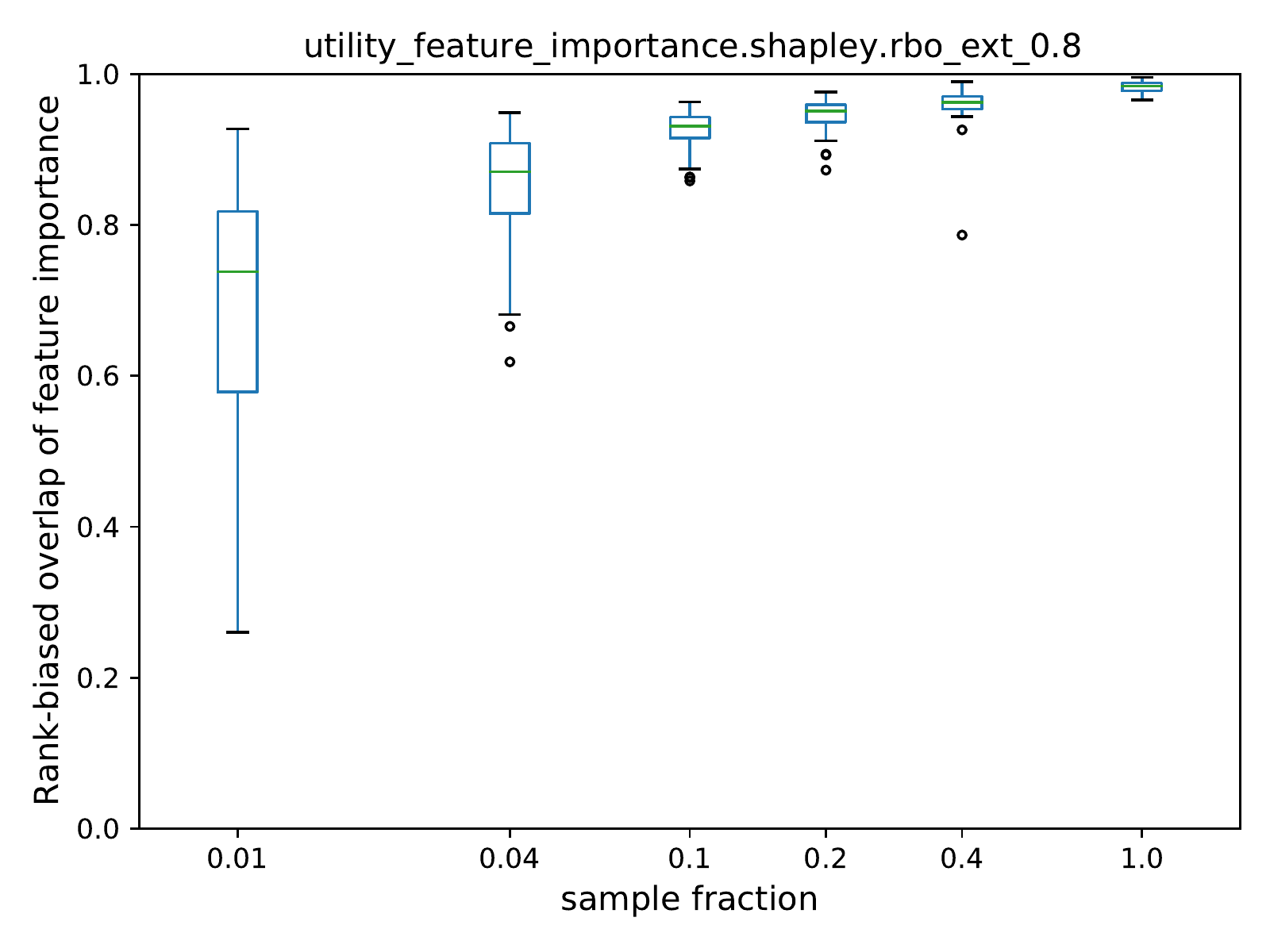}}}%
    \subfloat[\centering Household]{\includegraphics[width=0.35\columnwidth]{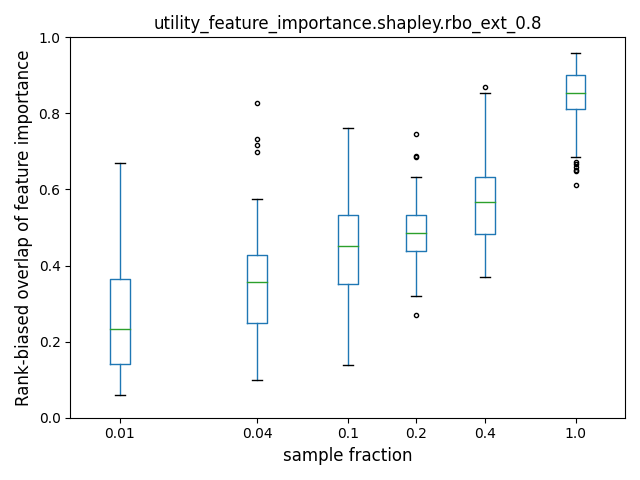}}%
    \subfloat[\centering Polish]{\includegraphics[width=0.35\columnwidth]{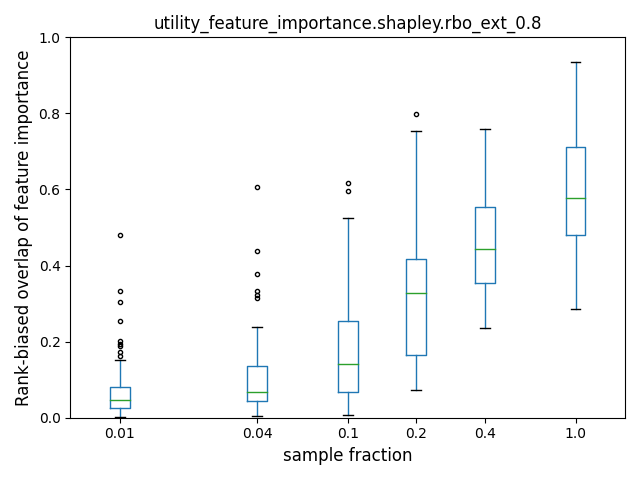}}%
    \caption{$\text{RBO}$ similarity between original dataset and different subsamples of the original dataset, ranging from a 1\% to a. 100\% fraction. The boxplots show the distribution of scores from 50 independent runs. The synthetic Adult dataset generally performs better vs. baselines compared to the other two datasets.}%
    \label{fig:subsample}%
\end{figure}

Figure \ref{fig:fs_real} shows the feature importance similarity between the original and the synthetic data, as measured by $\text{RBO}$, $\text{RBO}^{\text{Corr}}$ and Cosine similarity, for the three real-world datasets and the two different synthetic algorithms we used (PrivBayes - orange and PATE-CTGAN - blue). Each subplot shows how similarity changes with the epsilon bound chosen for the synthetic method. Each point represents the mean from 25 independent runs with the standard deviation also shown. The shaded areas represent the independent column resampling and random feature permutation baselines. In addition to the two baselines above, we also plot the $\text{RBO}$ measure between the original data and a random subsample of the original data (using resampling without replacement) in Figure~\ref{fig:subsample} for the Adult, Household and Polish datasets respectively.

The experiments show that similarity metrics take a range of values depending on the dataset, epsilon bound and synthetic method, with some patterns visible but a lot of case-specific variation. Some observations from Figure \ref{fig:fs_real} include: 
\begin{itemize}
    \item \textbf{Utility-privacy trade-off:} With PrivBayes there is a clear pattern of increasing similarity with epsilon, which is consistent with the findings of the experimental results on artificial data. It seems that PrivBayes allows a range of privacy-utility combinations and that the proposed measures capture this trade-off when it comes to utilising synthetic data for figuring out feature importance. Increasing the privacy level of the PrivBayes synthetic data leads to larger distortion in the quality of the features and the relationships between them, which leads to distortion in the feature importance scores and ranks. PATE-CTGAN's similarity values do not seem to vary consistently with $\epsilon$. Performance is typically stable with some changes in the amount of variance between runs.
    
    \item \textbf{Synthetic datasets are not always useful for performing feature engineering and measuring feature importance:} The results indicate that unless privacy guarantees are lowered, synthetic data cannot act as good proxies for original data when engineering features and analysing importances. For low and medium privacy bounds ($\epsilon < 0.4$), the feature similarity between PrivBayes and the original dataset is low for all datasets and similarity scores, and is lower or comparable to the two baseline synthesis methods. PATE-CTGAN similarity is more uniform across epsilon bounds and overlaps with the baselines in most cases, except for the Adult dataset where it is better. Its similarity scores have slightly lower variance than PrivBayes for the Household and Polish datasets. It is worth noting that epsilon values of 0.1 or 0.4 are not considered as extremely low or as offering very strong privacy [GM: Reference from literature here needed, e.g. Show me your epsilons]. The independent column resampling baseline performs better than both synthetic algorithms for low epsilon values, especially in the Household and Polish datasets. In addition, when comparing with the third baseline (see Figure~\ref{fig:subsample}, $\text{RBO}$ similarity achieved by PrivBayes and PATE-CTGAN with high epsilon bounds is comparable with that achieved by a 1\%-10\% fraction of the original dataset. With low epsilon bounds, similarity is lower than that of a 1\% subsample. A possible explanation for these results is that, with high privacy, the synthetic methods add so much noise to the data that the correlations between columns are effectively lost. Also, the marginal distributions of each column might become distorted as well, especially for continuous features which are discretised. Finally, there is generally large variance in all the similarity scores except when privacy is lowered significantly, demonstrating that synthetic data can lead to unpredictable results in contrast to the deterministic results produced by traditional anonymisation methods.
    
    In more detail, we observe the following patterns in individual datasets: Neither synthetic algorithm can replicate the hierarchical structure of the Household dataset, leading to assigning low scores to several important features, as shown previously. This affects the Cosine similarity a lot, as the angle between the importance vectors of the original and synthetic datasets increases. Synthetic Cosine similarity is almost never better than the independent column resampling baseline. Rank-based measures are also affected by this and overlap with the baselines. The Polish dataset performs only slightly better, with both synthetic algorithms overlapping significantly with the baselines and only performing better for high epsilon bounds. This is due to the uniformity of feature importances in this dataset, partially caused by multi-collinearity. Synthetic data perform better on the Adult dataset: Similarity measures are generally higher than the baselines for epsilon larger than 0.1 and PrivBayes has relatively small variance in this epsilon range. This is partially explained by the distinct importance values of several of the top features in the dataset, even though some of the less important features tend to have uniform importance (rank-based measures are affected much more by the leading features than the trailing ones). Also, the fact that Adult mostly consists of categorical variables and has a larger size than other datasets (which helps reduce variance between independent runs) might contribute to the higher similarity scores. 
    
    The use of synthetic data sometimes allows performing the preliminary task of ranking features when the dataset has specific characteristics or when privacy is compromised. But it is not accurate enough to allow these tasks in many of the real-world experimental scenarios that we tested, even though the datasets we are using are relatively small and with a limited number of features. This confirms some of the more recent cautionary tales found in the synthetic data literature \cite{oprisanu2021measuring}.
    
    \item \textbf{PrivBayes vs. PATE-CTGAN:} Comparing the two synthetic methods, PrivBayes is better than PATE-CTGAN for medium and high epsilon bounds, especially in the Adult dataset, but worse than PATE-CTGAN for low epsilon bounds. The proportionally larger number of categorical features in the Adult dataset might explain this variation.
    \item \textbf{Cautionary tale - Both synthetic algorithms have limited usability for more complex datasets with non-i.i.d. data and hierarchies:} Both synthetic methods perform poorly in the Household dataset, with much lower similarity values compared to other datasets. The reason this happens is that the HousehThe ataset contains a hierarchical structure which the two algorithms cannot replicate. As a result, features that summarise household members' information become much less useful and with lower scores as shown in the previous section and this is reflected in all three metrics. Performance for this dataset is comparable to the two baselines. In addition, PrivBayes similarity in this dataset increases weakly when decreasing privacy, showing that the main cause is due to the hierarchy and cannot be overcome with less noise.
    
    An additional advantage of PrivBayes is that the generative model it constructs (Bayesian network based on conditional distributions) is more explainable as it allows the user to know the estimated probability values of the joint distribution. In contrast, PATE-CTGAN is a black box which requires extra explainability tools to explain.
    
    \item \textbf{Larger datasets lead to smaller variance in metrics:} When comparing the Adult dataset with the Polish and Household datasets, it seems that the variance of all three metrics across independent runs is lower in the former (at least for PrivBayes). The reason for this is likely to be the larger size of this dataset compared to the other two. The same observation can be made in the subsample plots where larger fractions reduce variance.
    
    \item \textbf{$\text{RBO}^{\text{Corr}}$ captures more complete information about the relationships between variables:} There is a consistent increase in similarity when measuring it with $\text{RBO}^{\text{Corr}}$ compared to $\text{RBO}$. This is expected given that $\text{RBO}^{\text{Corr}}$ uses a broader definition of overlap and thus takes relationships between correlated variables into account. The increase is more marked in the Polish dataset, which could be due to its strong multi-collinearity. Apart from this increase in the  level, $\text{RBO}^{\text{Corr}}$ seems to generally follow the trend observed in $\text{RBO}$.
    
    \item \textbf{Rank-based and score-based metrics are both useful depending on the context:} Cosine similarity seems to have lower variance than the other two metrics. This might be due to the fact that Cosine similarity takes the feature scores into account, rather than just their ranks. Nevertheless, there are arguments in favour of using the two rank-based metrics alongside Cosine similarity; the scores generated by various well-known feature importance measures are not always reliable and all have well-documented issues and scenarios where they fail, while using the rankings and basing similarity in overlaps of feature lists might be less prone to those errors; $\text{RBO}$ and $\text{RBO}^{\text{Corr}}$ have the ability to give much more weight to top features, which is sometimes desirable (e.g. when the analyst is interested in selecting a small subset of features for their model or finding the top features for explainability purposes).
\end{itemize}

\subsection{Model Predictive Performance}

Model predictive performance: 
This is high and clearly higher than the resampling baseline. 
Is model training sometimes more robust than getting accurate feature importance? 

\begin{figure}[ht]
  \centering
 \includegraphics[width=1.0\columnwidth]{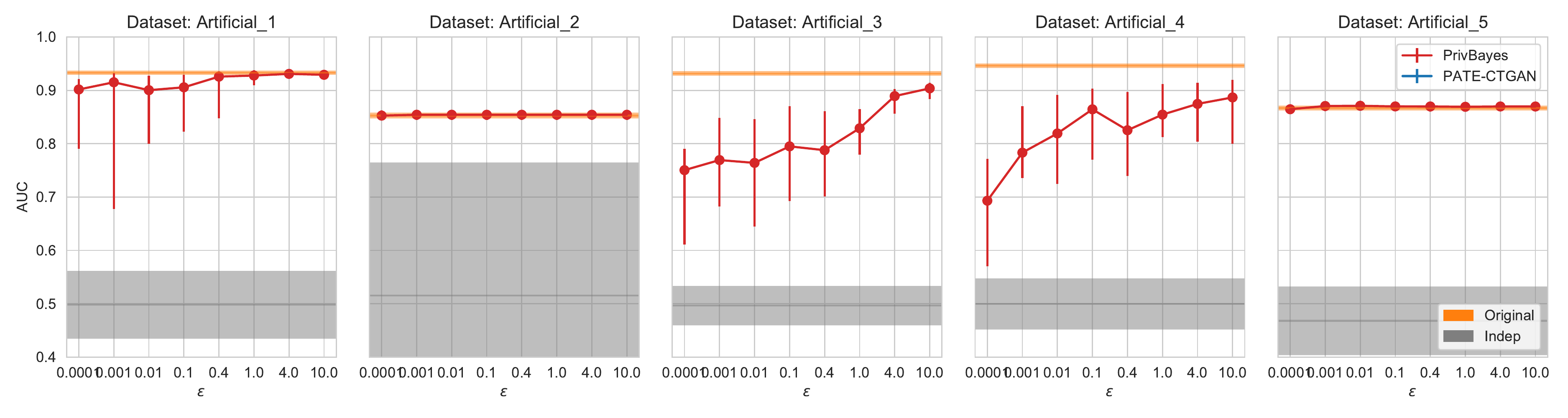}
    \caption{Artificial datasets 1-5: Model predictive performance as measured by AUC for the original datasets, PrivBayes synthetic datasets (for all epsilon values) and the independent column resampling dataset.}
    \label{fig:auc_artificial}
\end{figure}

\begin{figure}[ht]
  \centering
 \includegraphics[width=1.0\columnwidth]{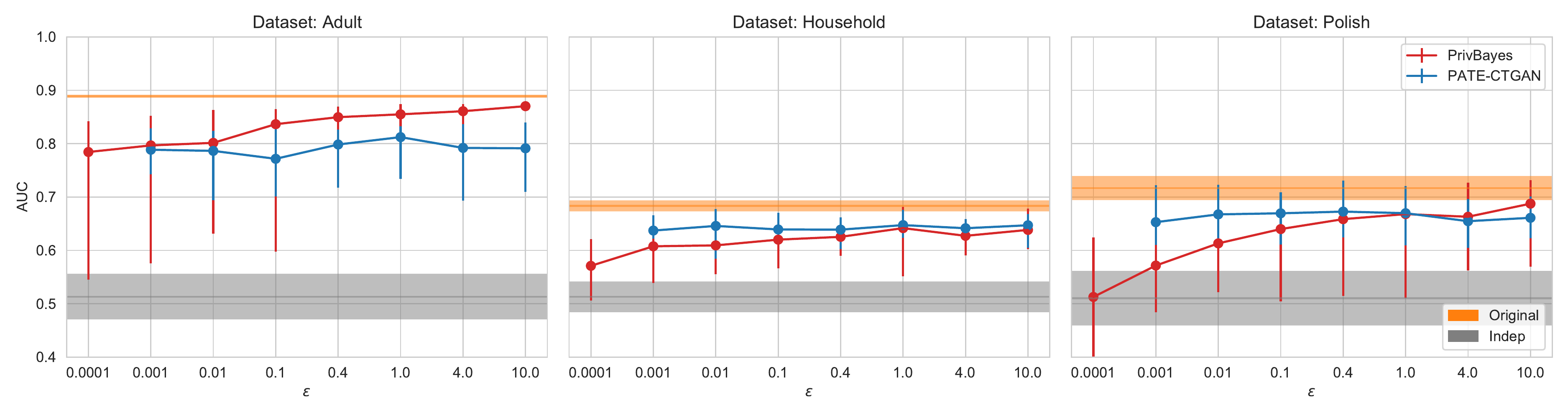}
    \caption{All real-world datasets: Model predictive performance as measured by AUC for the original datasets, synthetic datasets (for all epsilon values) and the independent column resampling dataset.}
    \label{fig:auc_real}
\end{figure}

Figures \ref{fig:auc_artificial} and \ref{fig:auc_real} show how the predictive performance of the Random Forest classifier changes with the epsilon bound, using the AUC metric. Artificial data runs only contain PrivBayes, while real-world dataset also include PATE-CTGAN. The figures show the performance of the original dataset, as well as the independent column resampling baseline. 

The artificial data results show that Artificial 3 and 4 perform worse than the other datasets, possibly because it is harder for the synthesisers to accurately capture all the relationships between variables when the number of variables increases. It is important to note that PrivBayes is set to use $k=3$, meaning that each variable is conditioned on a maximum of 3 other variables, which might be challenging for larger datasets. Larger values of $k$ lead to long computational times.

The real-world data results show that predictive performance is almost always higher than the independent column resampling baseline, except when using very low epsilon bounds and in the Polish dataset in particular. Performance is worse than when using the original data (and is close to the 0.5 random guess threshold in many cases) but gets close for high epsilon values. PrivBayes-generated datasets perform worse in the predictive task when epsilon is decreased, while PATE-CTGAN-generated datasets have approximately stable performance irrespective of epsilon. PrivBayes performs better in the Adult dataset and slightly worse in the other two datasets. 

The above results indicate that model predictive performance with synthetic data can at least outperform the baseline in the datasets we tested. This might be because machine learning models are able to train on imperfect data and still produce viable models, while feature importance (especially ranking features) can be more sensitive to the noise introduced by data synthesis, as described in the previous section. Still, model predictive performance with high privacy is low and noisy.

\section{Results and Conclusions}
\label{sec:conclusions}

Our empirical analysis of several artificial and real-world datasets indicates that there is no clear answer to the initial question of whether synthetic data can be used for performing early-stage tasks in the data science pipeline. Using the proposed similarity scores to quantify synthetic data utility in this usage scenario, we conclude that synthetic data do not produce accurate representations of feature importance when privacy is moderate or high ($epsilon > 0.1$), producing low mean similarity scores with large variance. Synthetic data similarity is often worse that the naive baselines we set up but becomes better when some privacy is compromised or when the dataset has characteristics that facilitate accurate data synthesis. 
    
We are able to draw some general conclusions about these characteristics from our experiments. Feature importance with synthetic data (and data synthesis in general) is easier (and with lower variance) when feature importances are not close to uniform and ideally when they are clearly distinct. Whether this happens depends on a number of factors and cannot always be predicted but we observe that datasets with high multicollinearity or features without much predictive value tend to make feature importances uniform and complicate the task. In addition, datasets with large numbers of rows and few columns are easier targets. Datasets with hierarchical structures (e.g. households and individuals) cannot be simulated accurately with the methods used in this paper and thus feature importance becomes inaccurate, especially if engineered features that summarise hierarchical information tend to be important. Finally, several synthetic methods are more suited to datasets with categorical features rather than numerical ones.
    
Using synthetic data for feature importance needs to be done with a lot of caution. We recommend that before synthetic data are used to perform any task, an initial investigation is warranted to understand the characteristics of the dataset in question and at least get a first estimate of whether synthetic methods are able to act as proxies. This investigation might involve preliminary runs in the original data, using some of the proposed metrics and algorithms (as well as others that might be relevant to particular applications). Unfortunately, this might defeat the purpose of starting preliminary analysis early, as this investigation requires some level of access to the original data or at least a sample of the original data.

Future work will involve experimenting with more complex datasets that are used in practice, applying some of the above techniques to understand the limits of using synthetic data to shorten project runtimes. Healthcare applications are of particular interest given the typically long and complicated processes to get access to sensitive data. Through these applications, it is possible to improve our understand about how different dataset characteristics (e.g. non-linearities) affect our ability to synthesise accurately and capture feature importance and whether other types of feature importance measures and similarity measures have advantages over the methods proposed here. Another stream of potential work consists of exploring additional utility metrics for feature importance as well as for other parts of preliminary data analysis like feature selection. In this work, we used differential privacy as a privacy indicator but there are multiple alternative ways to quantify privacy which are more interpretable, e.g. probability of identification after various types of intruder attacks. It would be interesting to understand how the utility-privacy trade-off we observed in this work shifts when using these alternative definitions of privacy. 


\bibliographystyle{ACM-Reference-Format}
\bibliography{refs}

\newpage
\appendix

\section{Appendix: Experimental set-up and process}
\label{subsec:setup}
In these experiments, we perform the following steps for each of the datasets, for each of the synthetic methods, and for each value of differential privacy budget, $\epsilon$, in $\{10^{-4}, 10^{-3}, 10^{-2}, 10^{-1}, 4 \cdot 10^{-1}, 1.0, 4.0, 10.0\}$:
\begin{enumerate}
    \item Train the synthetic model on the real dataset.
    \item Generate a synthetic dataset from the trained synthetic model.
    \item Perform automated feature engineering using the Featuretools Deep Feature Synthesis framework \citep{kanterDeepFeatureSynthesis2015} on the real and the synthetic dataset.
    \item Randomly split the real and synthetic datasets into and 70\% training sample and a 30\% validation sample.
    \item Independently for the real and synthetic datasets, train a Random Forest classifier on the training sample. The model predicts a selected variable in the dataset given all other variables.
    \item Calculate the three feature importance measures for the real and synthetic datasets, resulting in lists of feature rankings and scores.
    \item For each feature importance measure, calculate the four similarity measures between the real and synthetic datasets.
    \item Repeat steps (2)--(7) $l$ times with different random seeds.
    \item Repeat steps (1)--(8) $m$ times with different random seeds. 
\end{enumerate}

For PrivBayes, we set the parameter \emph{k} (maximum number of parent features) to 3 and $\beta$ to the default of 0.5, dividing the privacy budget evenly between estimating the network structure and the conditional probability distributions. For PATE-CTGAN we trained each model for approximately 3000 iterations so that the generated synthetic datasets, with different privacy budgets, are comparable. This is crucial when working with GAN-based models as the quality of synthetic datasets can vary significantly with the number of training iterations. 
For both PrivBayes and PATE-CTGAN we set $l=1$ and $m=25$.

When applying Featuretools to create engineered features, we made specific hyperparameter choices that fit each dataset and also allow us to explore how different types of feature engineering impact on the ability of synthetic methods to capture feature importance. For the Adult dataset, we configure Featuretools to create separate entities/tables for the variables ``education-num'', ``workclass'' and ``occupation'', to use ``max\_depth=2'' when creating features and we do not define specific primitives for feature creation. This results in a dataset with 141 features plus the predicted variable. For the Household dataset we configure Featuretools to create a separate entity containing all the household-level variables. We then create engineered features that summarise the information of all individuals in each household, using the primitives ``min'', ``max'', ``count'', ``mode'', ``num\_unique'', ``std'', ``sum'' and setting ``max\_depth=2''. This results in a dataset with 39 features plus the predicted variable. For the Polish dataset, we configure Featuretools to apply the transformation primitives ``multiply\_numeric'', ``subtract\_numeric'', ``add\_numeric'', ``divide\_numeric'', ``percentile'' to all numeric variables and to use ``max\_depth=2''. This results in a dataset with 67 features plus the predicted variable. 

When training the Random Forest classifiers, we predict the variable ``label'' for the Adult dataset, ``TenYearCHD'' for the Framingham dataset and ``Target'' for the Household dataset. We use the Scikit-learn \citep{scikit-learn} implementation of Random Forest and set the number of estimators in to 150. When calculating RBO and $RBO^{Cor}$ we set $p$ to 0.8.

We also implement three baselines for comparison:
\begin{enumerate}
    \item Random feature permutation: We randomly permute the feature rank list generated by a synthesis algorithm and then calculate similarity metrics. The performance of this baseline is expected to be at the lower end of similarity scores.
    \item Independent column resampling: This is an alternative synthesis algorithm that does not take into account the relationships between variables in the dataset. It resamples each column independently (with replacement) to generate a synthetic dataset. This is then used to calculate all the similarity metrics. This baseline is expected to result in a set of features with almost uniform importance, as they are all uncorrelated with the predicted variable. We use the Synthpop software package \citep{synthpop} to implement this baseline.
    \item Subsampling: We randomly subsample the rows of each original dataset and calculate the RBO similarity between the subsample and the original. Feature importance similarity is expected to drop with smaller subsamples.
\end{enumerate}

In addition to evaluating feature importance similarity, we also capture the predictive performance of the Random Forest classifier, using the area under the ROC Curve (AUC). For multi-class classification tasks (as we perform on the Household dataset) we calculate AUC using one-vs-one averaging, computing the average AUC of all possible pairwise combinations of classes \citep{auc_ovo}. We always train the classifiers on the 70\% training samples from the original and synthetic datasets. We always evaluate performance (AUC) on the 30\% validation sample taken from the original dataset (for both the original and synthetic datsets, AUC is calculated using the original hold-out sample).

All the code used in the experiments has been incorporated into the QUiPP synthetic data framework\footnote{\url{https://github.com/alan-turing-institute/QUIPP-pipeline}}, which contains a variety of synthetic data methods, utility and privacy metrics.




\end{document}